\documentclass[10pt,journal,compsoc]{IEEEtran}

\usepackage{graphicx}
\usepackage{amsmath}
\usepackage{amssymb}
\usepackage{arydshln}
\usepackage{epsfig}
\usepackage{multirow}
\usepackage{times}
\usepackage{array}
\usepackage{colortbl}
\usepackage{booktabs}
\usepackage{bbm}
\usepackage{multicol}
\usepackage{makecell}
\usepackage{xcolor}
\usepackage{xspace}
\usepackage{float}
\usepackage{color}
\usepackage{stfloats}
\usepackage{mathtools}
\usepackage{textcomp}
\usepackage{gensymb}
\usepackage{siunitx}
\usepackage{algorithmic}
\usepackage{algorithm}
\usepackage{listings}
\usepackage{boldline}
\usepackage{lipsum}
\usepackage{setspace}
\usepackage{caption}
\usepackage{flushend,cuted}
\usepackage{xtab,booktabs}
\usepackage{balance}
\usepackage{hyperref}
\hypersetup{
	colorlinks=true,
	linkcolor=black,
	filecolor=blue,      
	urlcolor=black,
	citecolor=green,
    breaklinks=true
}
\usepackage{cite}

\usepackage{multicol}
\usepackage{lipsum}

\newcommand{\wugetable}[1]{{\textcolor{black}{#1}}} 

\newcommand{\ylfChecked}[1]{{\textcolor{black}{#1}}}

\ifCLASSINFOpdf
\else
\fi

\begin{document}
%
\title{A Survey of Historical Learning:\\ Learning Models with Learning History}
%
%
%
%
\author{
        Xiang Li$^*$,
	    Ge Wu$^*$,
	    Lingfeng Yang,
        Wenhai Wang,
        Renjie Song,
		Jian Yang$^\#$\thanks{$*$ Equal contribution. $\#$ Corresponding author.}
		\IEEEcompsocitemizethanks{
		    
		    \IEEEcompsocthanksitem X. Li, G. Wu and J. Yang are from IMPlus@PCALab, College of Computer Science, Nankai University. Email: xiang.li.implus@nankai.edu.cn, gewu.nku@gmail.com, csjyang@nankai.edu.cn
		    \IEEEcompsocthanksitem L. Yang is with PCALab, School of Computer Science and Engineering, Nanjing University of Science and Technology. Email: yanglfnjust@njust.edu.cn
            \IEEEcompsocthanksitem W. Wang is with Shanghai AI Laboratory. Email: wangwenhai362@gmail.com
            \IEEEcompsocthanksitem R. Song is from Megvii Technology, Nanjing. Email: songrenjie@megvii.com
		}
	}

%
%

\markboth{}
{Shell \MakeLowercase{\textit{et al.}}: Bare Demo of IEEEtran.cls for Computer Society Journals}
%



\IEEEtitleabstractindextext{%

\begin{abstract}
New knowledge originates from the old. The various types of elements, deposited in the training history, are a large amount of wealth for improving learning deep models. In this survey, we comprehensively review and summarize the topic--``Historical Learning: Learning Models with Learning History'', which learns better neural models with the help of their learning history during its optimization, from three detailed aspects: Historical Type (what), Functional Part (where) and Storage Form (how). To our best knowledge, it is the first survey that systematically studies the methodologies which make use of various historical statistics when training deep neural networks. The discussions with related topics like recurrent/memory networks, ensemble learning, and reinforcement learning are demonstrated. We also expose future challenges of this topic and encourage the community to pay attention to the think of historical learning principles when designing algorithms. The paper list related to historical learning is available at \url{https://github.com/Martinser/Awesome-Historical-Learning.}

\end{abstract}

}

\maketitle

\begin{figure*}[t]
	\vspace{0pt}
	\begin{center}
		\setlength{\fboxrule}{0pt}
		\fbox{\includegraphics[width=0.98\textwidth]{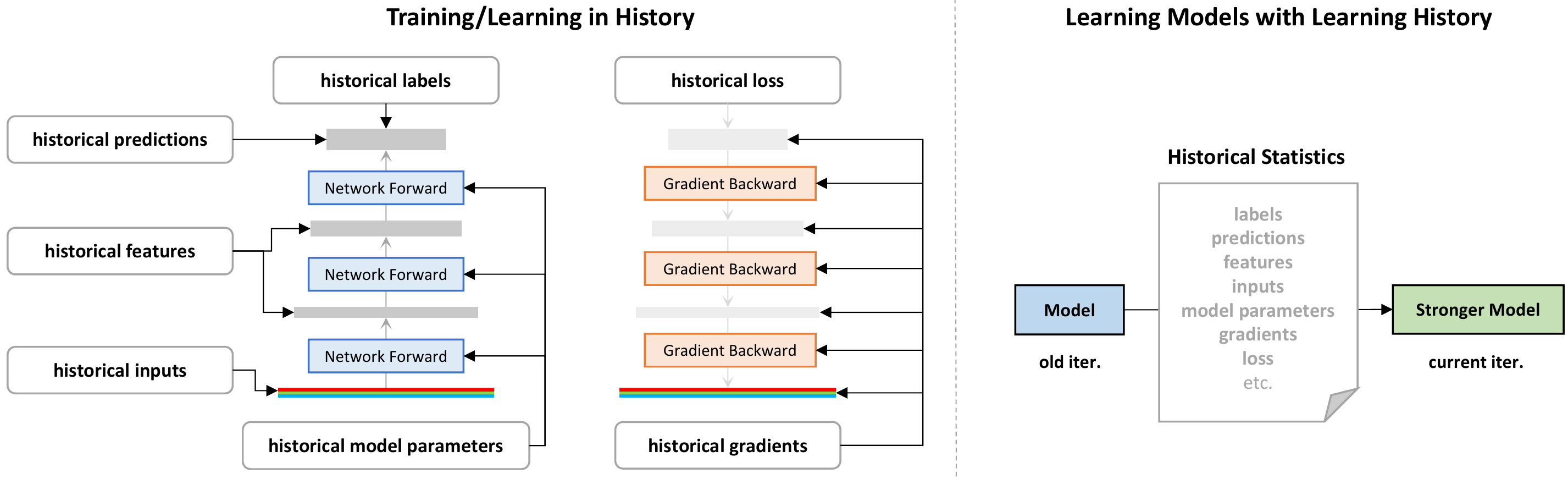}}
	\end{center}	
	\vspace{-10pt}
	\caption{ 
	\textbf{The concept of ``Learning Models with Learning History''.} During optimization, deep models can be improved potentially by exploiting the historical statistics, including the past labels, predictions, feature representations, inputs, model parameters, gradients, losses, and their combinations. ``iter.'' denotes iteration.
	}
	\label{fig_hl_concept}
\end{figure*}

\IEEEdisplaynontitleabstractindextext

%
\IEEEpeerreviewmaketitle

\section{Introduction}
\noindent\emph{``He who by reviewing the old can gain knowledge of the new and is fit to be a teacher.''}--Confucius, China.


{People gain new knowledge and make advances by continuously valuing their past experiences. These varied experiences may include hard-won lessons, innovative perspectives, and valuable introspection on certain topics. This wealth of knowledge allows individuals to make ongoing progress and become experts in their fields.}


{This phenomenon is also present in machine intelligence. When deep learning models are trained, they accumulate a large amount of diverse historical data, including gradients, model parameters, predictions, intermediate feature representations, labels, inputs, and losses. Just like in the human system, the accumulation of past information during the learning process has the potential to significantly enhance the intelligence of these models.}


{In this paper, we summarize and formulate the topic of ``Historical Learning: Learning Models with Learning History'' (Fig.~\ref{fig_hl_concept}). Our goal is to provide a comprehensive and systematic view of using historical elements during optimization and to explore the potential for utilizing this diverse knowledge to benefit machine learning models. We aim to answer the questions of what, where, and how this information can be exploited. Additionally, we discuss the relationships between historical learning and other related topics and outline the challenges that must be addressed in order to effectively use historical data to improve learning systems.
}

{There have been a number of previous works that explore the use of historical data in machine learning models. These methods typically focus on five types of historical data:} 
(1) \textbf{Gradients}. All the optimizers~\cite{liu2019variance,loshchilov2017decoupled,kingma2014adam,sutskever2013importance,zeiler2012adadelta,duchi2011adaptive, dozat2016incorporating, chen2018closing, heo2020adamp, xie2022adan,ding2019adaptive} using momentum-based operators have made good use of the past gradients of model parameters to improve the training convergence. Equalization loss~\cite{tan2021equalization,li2022equalized} and~\cite{kwon2020backpropagated} adopt the historical gradient-guided mechanism in long-tailed training or abnormal sample detection. 
(2) \textbf{Model parameters}. The historical architectures are generally utilized as discrete ensemble checkpoints~\cite{huang2017snapshot,chen2017checkpoint,chen2018short,ilg2018uncertainty,poggi2020uncertainty,wolterink2018automatic,yu2019deep,liu2020loss,xu2020multi,gong2022uncertainty,zhang2020swa,athiwaratkun2018there,izmailov2018averaging,yang2019swalp,cha2021swad,guo2022stochastic,garipov2018loss,maddox2019simple,von2020neural} for inference or as moving-averaged/direct teachers~\cite{caron2021emerging,chen2021empirical,feng2021temporal,grill2020bootstrap,he2020momentum,chen2020improved,el2021training,chen2021transferrable,luo2021coco,chen2020big,van2021unsupervised,xie2021propagate,li2020prototypical,ayush2021geography,xie2021detco,tomasev2022pushing,deng2021unbiased,cai2019exploring,
mitrovic2020representation,yuan2020revisiting,li2019learning,yang2019snapshot,furlanello2018born,wang2022efficient,wang2022learn,wang2022self,tarvainen2017mean,cai2021exponential, zhang2023robust,
cai2022semi,wang2022unsupervised,shi2018transductive,wang2021multi, kim2022pose, yu2019uncertainty, meng2021spatial, wang2021self,liu2022perturbed, cui2019semi, perone2018deep,zhang2022semi, xu2021end, yang2021interactive , liu2021unbiased, liu2022unbiased,wang2022semi} to provide extra meaningful supervisions. 
(3) \textbf{Feature representations}. The stored old feature maps efficiently augment data points to construct supplementary/critical training pairs~\cite{li2021spatial,zhuang2019local,zhou2022regional,zheng2021group,sprechmann2018memory,wang2020cross,deng2022insclr,yang2022online,deng2021variational,li2020unsupervised,feng2021exploring,liu2022memory,xiao2017joint,alonso2021semi,lee2021weakly,li2022recurrent,liu2021one,yang2018learning,yang2019visual,el2021training,liu2022densely,wang2022contrastive,wen2021false,liu2022learning,liu2021noise,ortego2021multi,jin2021mining,yang2022cross,zhong2019invariance,wu2018improving,wu2018unsupervised,misra2020self, tian2020contrastive, yin2021instance, dai2021dual,wang2022semi,ko2021learning,he2020momentum, chen2020improved,chen2021transferrable,luo2021coco,chen2020big,van2021unsupervised,xie2021propagate,li2020prototypical,ayush2021geography, xie2021detco
} or class centers~\cite{wen2016discriminative,he2018triplet,li2019angular,min2020adversarial,zhao2020deep,narayan20193c,chen2021free,li2021frequency, keshari2020generalized, jing2021cross,fan2018associating,liu2020leveraging}. Their statistics (e.g., mean and variance) are sometimes accumulated for better training stability/convergence~\cite{ioffe2015batch,caron2021emerging,ioffe2017batch, li2016revisiting, salimans2016weight, arpit2016normalization}. 
(4) \textbf{Probability predictions}. The recorded probability predictions of specific instances are adopted for ensembling~\cite{laine2016temporal,yang2022recursivemix, yao2020deep,chen2020semi} and self-teaching~\cite{kim2021self,nguyen2019self, shen2022self}. 
(5) \textbf{Loss values}. The statistics of loss values are used to identify
anomalous data~\cite{chu2020neural} and biased data~\cite{jiang2021delving,huang2019o2u,xu2021small,yue2022ctrl,lazarou2021iterative,xia2021sample}. 
{In addition to these historical types, we also consider the functional part (where the data is used) and the storage form (how it is stored) of historical statistics to better understand how it is applied in these methods. These existing works provide strong evidence that historical statistic is important for training more effective deep learning models.}

{To the best of our knowledge, we are the first to provide a systematic review and formulation of the topic ``Historical Learning: Learning Models with Learning History.'' We thoroughly summarize and attribute the existing representative works in this field, examining them from three detailed aspects. We also discuss the relationships between this topic and several related areas and outline the challenges that will encourage the community to continue exploring this topic in the future.}


\section{Learning Models with Learning History}
In this section, we summarize and describe the concept of 
{``Historical Learning: Learning Models with Learning History''}
in details.

Assume a deep model has $n$ hierarchical layers with total $\{W_1, ...,  W_n\}$ parameters, and it takes input $x_0$ to produce the prediction $p$, i.e., $p=\sigma(x_n)$, where $x_n=f(x_0, \{W_1, ...,  W_n\})$ and $\sigma$ is the output activation function. The loss $l$ is obtained from $p$ under the current supervision $y$. The feature $x_i$ generated by each layer $i\ (0 < i \le n)$ constitutes the intermediate representations $\{x_1, ...,  x_n\}$ at all scales. Based on the chain rule of backpropagation~\cite{lecun1988theoretical}, we have two groups of the gradients calculated on loss $l$: $\{\nabla_{W_1}, ..., \nabla_{W_n}\}$ w.r.t. the parameters and $\{\nabla_{x_0}, ..., \nabla_{x_n}\}$ w.r.t. the input and all-level features maps. Specifically, we use the superscript $t$ (e.g., $p^t$) to mark the moment of each variable in the training process. The superscript $t$ can be epoch-wise, iteration-wise, or customized period-wise.

Given moment $t$, the historical information $H$ is then defined as a superset of all the past statistics: 
\begin{equation}
\begin{aligned}
	\vspace{0pt}
    H = \Big\{ x_0^j, y^j, p^j, l^j,  \{x_1^j, ...,  x_n^j \},  \{W_1^j, ...,  W_n^j\}, \\ \{\nabla_{x_0}^j, ..., \nabla_{x_n}^j\}, \{\nabla_{W_1}^j, ..., \nabla_{W_n}^j\} |  \forall j < t \Big\}.
    \label{eqn_history}
\end{aligned}
\end{equation}

\ylfChecked{Any algorithm $\mathcal{A}(\cdot)$ that uses the statistical information from the defined history $H$ (i.e., $\mathcal{A}(H)$) to improve model performance can be considered ``Learning Models with Learning History.'' Meanwhile, the historical statistical information can be categorized into three primary elements - data, model, and loss~\cite{lecun2015deep} - based on their functional role in the typical process of training deep learning models.}

\ylfChecked{It is important to note that the entire history $H$ can occupy a substantial amount of storage space, and it may be impractical or inefficient to store all of it on GPU or CPU devices. To address this issue, there are two commonly used methods in the literature: (1) Discrete Elements (DE), which entails storing a limited number of historical checkpoints for later use, and (2) Moving Average (MA), which involves applying momentum or summary updates to the stored variables.}

{In summary, the topic of ``Historical Learning: Learning Models with Learning History'' involves using historical data related to the model during its optimization in the form of Discrete Elements (DE) or Moving Average (MA), to improve the generalization and performance of the model. In the next section, we provide a detailed review of existing approaches to this topic from the three perspectives mentioned above.}

\section{Review of Existing Approaches}
\begin{figure}[htp]
	\vspace{0pt}
		\setlength{\fboxrule}{0pt}
\fbox{\includegraphics[width=0.5\textwidth,height=0.9\textheight,keepaspectratio] 
{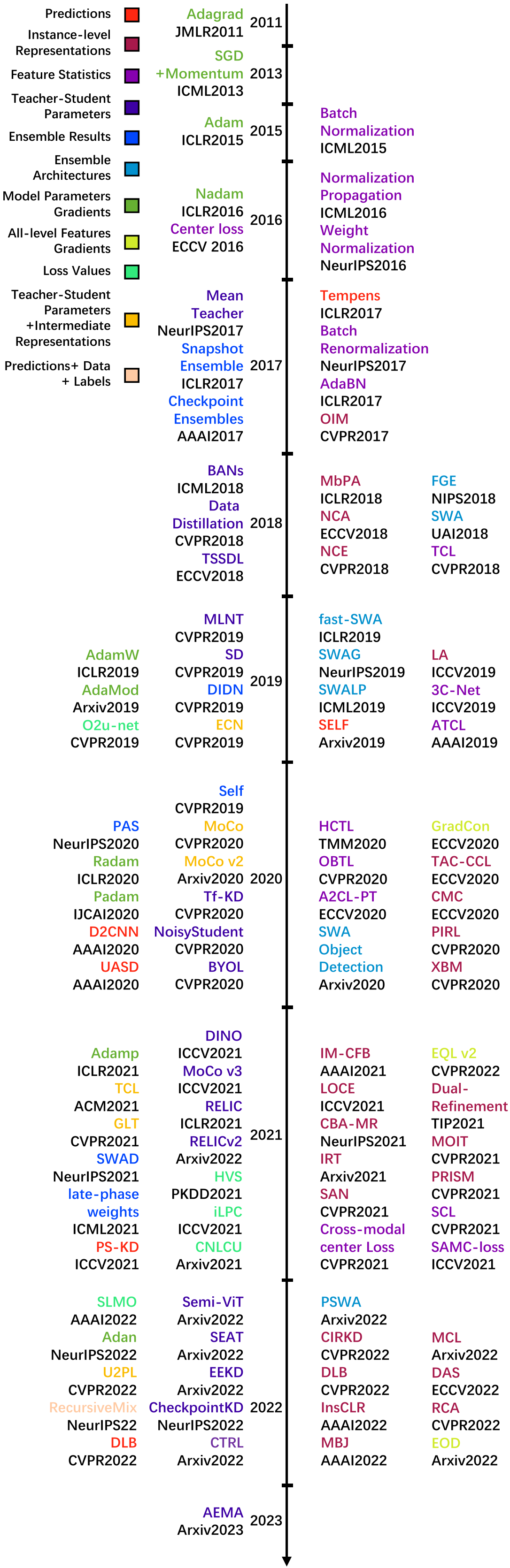}}
	\vspace{0pt}
	\caption{\textbf{The timeline of the works related to historical learning}. Different colors represent different historical information or types involved. For example, the red color is related to past predictions, while the light orange involves historical predictions, data, and labels. It demonstrates the tendency of using different historical types.}
	\label{fig_timeline}
	\vspace{0pt}
\end{figure}

Table~\ref{table1} summarizes the representative works related to ``Learning Models with Learning History'', and the timeline of these works is shown in Fig.~\ref{fig_timeline}. We elaborate on their details in the following subsections from multiple aspects.  

\subsection{Aspect of Historical Type~\label{sec_3.1}}
\noindent\textbf{Prediction.} It is known that the past predictions of specific instances are from architectures with diverse parameters and distinct augmented inputs. Therefore, the prediction can carry informative and complementary probability distributions, making it potential to be a teacher that can guide the learning of the original model. {Following this intuition, Kim et al.~\cite{kim2021self} adopt the prediction only from the last epoch. Thus, the new target~$p^{t}$ of PS-KD at epoch $t$ can be written as:
\begin{equation}
\begin{aligned}
p^{t}=\left(1-\alpha^{t}\right) {y}^t+\alpha^{t} p^{t-1},
\end{aligned}
\label{eq_target}
\end{equation}
\begin{equation}
\begin{aligned}
\alpha^{t}=\alpha^{T} \times \frac{t}{T}.
\label{eq_alpha}
\end{aligned}
\end{equation}
Eq.~\eqref{eq_target} represents that the current soft target~$p^{t}$ is a combination of the past~$(t-1)$-th epoch prediction~$p^{t-1}$ and the one-hot target~$y$. Moreover, the hyperparameter~$\alpha$ changes continuously with epoch $t$ in Eq.~\eqref{eq_alpha}. Through the use of historical predictions, the model can implicitly change the weight of hard samples, thereby paying more attention to these samples. {Chen et al.~\cite{chen2020semi} also integrate historical predictions to produce a smoother adaptive distribution for the current iteration. Besides, employing historical predictions has been effectively implemented in the temporal ensembling method by Laine et al.~\cite{laine2016temporal},}
\ylfChecked{which uses the Exponential Moving Average (EMA) to update predictions among all past epochs:}
\begin{equation}
\begin{aligned}
p^t  =\alpha p^{t-1}+(1-\alpha) p ^t.
\end{aligned}
\end{equation}
The prediction of the network~$p^t$ is cumulatively calculated with the historical prediction~$p^{t-1}$. Additionally,~$\alpha$~represents the momentum term, which determines the proportion of historical prediction participating in the $t$-th iteration. Shen et al.~\cite{shen2022self} use half of each mini-batch for extracting smoothed labels from previous iterations and the other half for providing soft targets for self-regularization.}

{Motivated by the work in~\cite{laine2016temporal}, some researchers have used temporal ensembling to improve network performance. Yao et al.~\cite{yao2020deep} use historical predictions of previous epochs to obtain more accurate confidence levels for different labels. Nguyen et al.~\cite{nguyen2019self} solve the issue of network overfitting in noisy datasets by filtering out incorrect labels during training and using EMA to ensemble historical predictions. Different from above work, Yang et al.~\cite{yang2022recursivemix} propose RecursiveMix, which takes the prediction from historical input as the target and aligns it with the semantics of the corresponding ROI area in the current training input.}

\begin{strip}
\centering
\phantomsection{\label{table_review}}
\footnotesize
\setlength{\tabcolsep}{4.pt}
\vspace{0pt}
\renewcommand\arraystretch{1.2}
\topcaption{\label{table1}Review of the representative works which ``learn models with learning history'' from the aspects of Historical Type (what), Functional Part (where), and Storage Form (how). }
\vspace{-10pt}
\begin{xtabular}{l|l||c|c|c|c|c|c|c|c||c|c|c||c|c}
        \hline
        \multirow{2}{*}{Reference} & \multirow{2}{*}{Method} & \multicolumn{8}{c||}{Historical Type}                    & \multicolumn{3}{c||}{Functional Part}     & \multicolumn{2}{c}{Storage Form} \\
        \cline{3-15}
        &                                                    & $p$ & $\{x_i\}$ & $x_0$ & $y$ & $\{W_i\}$ & $\{\nabla_{x_i}\}$ & $\{\nabla_{W_i}\}$ & $l$     & Data & Model & Loss                       &\ \ DE\ \ & MA  \\
        \hline
        \wugetable{CVPR22} & Shen et al.~\cite{shen2022self}  &  $\checkmark$ & &  &  &  & & & & & & $\checkmark$ &    $\checkmark$ &   \\ 
        ICCV21 & Kim et al.~\cite{kim2021self}  &  $\checkmark$ & &  &  &  & & & & & & $\checkmark$ &    $\checkmark$ &   \\ 
        \wugetable{AAAI20} & Chen et al.~\cite{chen2020semi} &  $\checkmark$ & &  &  &  & & & & & & $\checkmark$ &  &  $\checkmark$    \\ 
        \wugetable{AAAI20} & Yao et al.~\cite{yao2020deep} & $\checkmark$ &  &  & &   & & & &  & & $\checkmark$ &  &  $\checkmark$  \\ 
        \wugetable{Arxiv19} & Nguyen et al.~\cite{nguyen2019self} & $\checkmark$ &  &  & &   & & & &  & & $\checkmark$ &  &  $\checkmark$  \\ 
        ICLR17 & Laine et al.~\cite{laine2016temporal} & $\checkmark$ &  &  & &   & & & &  & & $\checkmark$ &  &  $\checkmark$  \\ 
        \hline
        \wugetable {AAAI22} & Liu et al.~\cite{liu2022memory} &  & $\checkmark$ &  & &  &  & & & $\checkmark$ & & $\checkmark$ &  $\checkmark$ &   \\
        \wugetable {AAAI22 }& Wang et al.~\cite{wang2022contrastive} &  & $\checkmark$ &  & &  &  & & & $\checkmark$ & & $\checkmark$ &  $\checkmark$ &   \\ 
        \wugetable {AAAI22} & Deng et al.~\cite{deng2022insclr} &  & $\checkmark$ &  & &  &  & & & $\checkmark$ & & $\checkmark$ &  $\checkmark$ &  \\ 
        \wugetable{ CVPR22}&  Zhou et al.~\cite{zhou2022regional} &  & $\checkmark$ &  & &  &  & & & $\checkmark$ & & $\checkmark$ &  $\checkmark$ &  $\checkmark$\\
        \wugetable{ CVPR22}& Liu et al.~\cite{liu2022learning} &  & $\checkmark$ &  & &  &  & & & $\checkmark$ & & $\checkmark$ &  $\checkmark$ &  \\
       \wugetable{CVPR22} & Yang et al.~\cite{yang2022cross} &  & $\checkmark$ &  & &  &  & & & $\checkmark$ & & $\checkmark$ &  $\checkmark$ &  \\ 
       \wugetable{CVPR22} & Li et al.~\cite{li2022recurrent} &  & $\checkmark$ &  & &  &  & & & $\checkmark$ & & $\checkmark$ &  $\checkmark$ &  \\ 
       \wugetable{ ECCV22} & Liu et al.~\cite{liu2022densely} &  & $\checkmark$ &  & &  &  & & & $\checkmark$ & & $\checkmark$ &  $\checkmark$ &  \\
       \wugetable {Arxiv22}& Yang et al.~\cite{yang2022online} &  & $\checkmark$ &  & &  &  & & & $\checkmark$ & & $\checkmark$ &  $\checkmark$ &  \\
       \wugetable {AAAI21} & Yin et al.~\cite{yin2021instance} &  & $\checkmark$ &  & &  &  & & & $\checkmark$ & & $\checkmark$ &  $\checkmark$ &  \\
       \wugetable{CVPR21} & Zheng et al.~\cite{zheng2021group} &  & $\checkmark$ &  & &  &  & & & $\checkmark$ & & $\checkmark$ &  $\checkmark$ & $\checkmark$  \\ 
       \wugetable{ CVPR21} & Liu et al.~\cite{liu2021one} &  & $\checkmark$ &  & &  &  & & & $\checkmark$ & & $\checkmark$ &  $\checkmark$ &  \\
        \wugetable{ CVPR21} & Feng et al.~\cite{feng2021exploring} &  & $\checkmark$ &  & &  &  & & & $\checkmark$ & & $\checkmark$ &  $\checkmark$ &  \\
        \wugetable{ CVPR21} & Deng et al.~\cite{deng2021variational} &  & $\checkmark$ &  & &  &  & & & $\checkmark$ & & $\checkmark$ &  $\checkmark$ &  \\
        \wugetable{ CVPR21} & Jin et al.~\cite{jin2021mining} &  & $\checkmark$ &  & &  &  & & & $\checkmark$ & & $\checkmark$ &  $\checkmark$ &  \\ 
        \wugetable{CVPR21} & Li et al.~\cite{li2021spatial}&  & $\checkmark$ &  & &  &  & & & $\checkmark$ & & $\checkmark$ &  $\checkmark$ & $\checkmark$ \\ 
        \wugetable{CVPR21} & Liu et al.~\cite{liu2021noise} &  & $\checkmark$ &  & &  &  & & & $\checkmark$ & & $\checkmark$ &  $\checkmark$ &  \\ 
        \wugetable{ICCV21} & Ko et al.~\cite{ko2021learning}&  & $\checkmark$ &  & &  &  & & & $\checkmark$ & & $\checkmark$ &  $\checkmark$ &   \\ 
        \wugetable{ ICCV21} & Alonso et al.~\cite{alonso2021semi} &  & $\checkmark$ &  & &  &  & & & $\checkmark$ & & $\checkmark$ &  $\checkmark$ &  \\
       \wugetable{ ICCV21} & Lee et al.~\cite{lee2021weakly} &  & $\checkmark$ &  & &  &  & & & $\checkmark$ & & $\checkmark$ &  $\checkmark$ &  \\
       \wugetable{ICCV21} & Feng et al.~\cite{feng2021exploring} &  & $\checkmark$ &  & &  &  & & & $\checkmark$ & & $\checkmark$ &  $\checkmark$ &  \\ 
        \wugetable{ NeurIPS21} & Wen et al.~\cite{wen2021false} &  & $\checkmark$ &  & &  &  & & & $\checkmark$ & & $\checkmark$ &  $\checkmark$ &  \\
        \wugetable{TIP21} & Dai et al.~\cite{dai2021dual} &  & $\checkmark$ &  & &  &  & & & $\checkmark$ & & $\checkmark$ &  $\checkmark$ &  \\ 
        CVPR20 & Wang et al.~\cite{wang2020cross} &  & $\checkmark$ &  & &  &  & & & $\checkmark$ & & $\checkmark$ &  $\checkmark$ &  \\ 
        \wugetable{CVPR20} &Misra et al.~\cite{misra2020self} &  & $\checkmark$ &  & &  &  & & & $\checkmark$ & & $\checkmark$ &  $\checkmark$ &  \\ 
        \wugetable{ECCV20} & Li et al.~\cite{li2020unsupervised} &  & $\checkmark$ &  & &  &  & & & $\checkmark$ & & $\checkmark$ &  $\checkmark$ &  \\ 
        \wugetable{ECCV20} & Tian et al.~\cite{tian2020contrastive} &  & $\checkmark$ &  & &  &  & & & $\checkmark$ & & $\checkmark$ &  $\checkmark$ &  \\ 
        CVPR19 & Zhong et al.~\cite{zhong2019invariance} &  & $\checkmark$ &  & &  & & &  &  &  & $\checkmark$ &  & $\checkmark$ \\  
        \wugetable{ICCV19} & Zhuang et al.~\cite{zhuang2019local}  &  & $\checkmark$ &  & &  &  & & & $\checkmark$ & & $\checkmark$ &  $\checkmark$ &$\checkmark$  \\
         &                                                    & $p$ & $\{x_i\}$ & $x_0$ & $y$ & $\{W_i\}$ & $\{\nabla_{x_i}\}$ & $\{\nabla_{W_i}\}$ & $l$     & Data & Model & Loss                       &\ \ DE\ \ & MA  \\\hline
        \wugetable{TPAMI19} & Yang et al.~\cite{yang2019visual} &  & $\checkmark$ &  & &  &  & & & $\checkmark$ & & $\checkmark$ &  $\checkmark$ &  \\ 
        CVPR18 & Wu et al.~\cite{wu2018unsupervised}  &  & $\checkmark$ &  & & &  & & & $\checkmark$ &  & $\checkmark$ & $\checkmark$ &  \\  
        ECCV18 & Wu et al.~\cite{wu2018improving} &  & $\checkmark$  &   & & & & & & $\checkmark$ &  & $\checkmark$ &   & $\checkmark$  \\ 
        \wugetable{ECCV18} & Yang et al.~\cite{yang2018learning} &  & $\checkmark$ &  & &  &  & & & $\checkmark$ & & $\checkmark$ &  $\checkmark$ &  \\ 
        \wugetable{ICLR18} & Sprechmann et al.~\cite{sprechmann2018memory}  &  & $\checkmark$ &  & &  &  & & & $\checkmark$ & & $\checkmark$ &  $\checkmark$ &   \\  
        \wugetable{CVPR17} & Xiao et al.~\cite{xiao2017joint} &  & $\checkmark$ &  & &  &  & & & $\checkmark$ & & $\checkmark$ &  $\checkmark$ &  \\ 
        \wugetable{CVPR21}  & Li et al.~\cite{li2021frequency} &  & $\checkmark$ &  & &  & & & & $\checkmark$  &  & $\checkmark$ &  & $\checkmark$   \\ 
        \wugetable{CVPR21}  & Jing et al.~\cite{jing2021cross} &  & $\checkmark$ &  & &  & & & & $\checkmark$  &  & $\checkmark$ &  & $\checkmark$   \\ 
         \wugetable{ICCV21 } & Chen et al.~\cite{chen2021free} &  & $\checkmark$ &  & &  & & & & $\checkmark$  &  & $\checkmark$ &  & $\checkmark$   \\ 
        \wugetable{ CVPR20}  & Keshari et al.~\cite{keshari2020generalized} &  & $\checkmark$ &  & &  & & & & $\checkmark$  &  & $\checkmark$ &  & $\checkmark$   \\ 
        \wugetable{ECCV20}  & Min et al.~\cite{min2020adversarial} &  & $\checkmark$ &  & &  & & & & $\checkmark$  &  & $\checkmark$ &  & $\checkmark$   \\ 
        \wugetable{TPAMI20}  & Liu et al.~\cite{liu2020leveraging} &  & $\checkmark$ &  & &  & & & & $\checkmark$  &  & $\checkmark$ &  & $\checkmark$   \\
        \wugetable{TMM20}  & Zhao et al.~\cite{zhao2020deep} &  & $\checkmark$ &  & &  & & & & $\checkmark$  &  & $\checkmark$ &  & $\checkmark$   \\ 
        \wugetable{ AAAI19} & Li et al.~\cite{li2019angular} &  & $\checkmark$ &  & &  & & & & $\checkmark$  &  & $\checkmark$ &  & $\checkmark$   \\ 
       \wugetable{ ICCV19}  & Narayan et al.~\cite{narayan20193c} &  & $\checkmark$ &  & &  & & & & $\checkmark$  &  & $\checkmark$ &  & $\checkmark$   \\
        \wugetable{CVPR18} & He et al.~\cite{he2018triplet} &  & $\checkmark$ &  & &  & & & & $\checkmark$  &  & $\checkmark$ &  & $\checkmark$   \\ 
        \wugetable{ECCV18} & Fan et al.~\cite{fan2018associating} &  & $\checkmark$ &  & &  & & & & $\checkmark$  &  & $\checkmark$ &  & $\checkmark$   \\
        ECCV16 & Wen et al.~\cite{wen2016discriminative} &  & $\checkmark$ &  & &  & & & & $\checkmark$  &  & $\checkmark$ &  & $\checkmark$   \\
        \hline  
        \wugetable{ICLR17} & Li et al.~\cite{li2016revisiting} &  & $\checkmark$  &  & &   & & & &  & $\checkmark$ & & &  $\checkmark$  \\
        \wugetable{NeurIP17} & Ioffe et al.~\cite{ioffe2017batch} &  & $\checkmark$  &  & &   & & & &  & $\checkmark$ & & &  $\checkmark$  \\     
        \wugetable{ICML16} & Arpit et al.~\cite{arpit2016normalization} &  & $\checkmark$  &  & &   & & & &  & $\checkmark$ & & &  $\checkmark$  \\ 
        \wugetable{NeurIP16} & Salimans et al.~\cite{salimans2016weight} &  & $\checkmark$  &  & &   & & & &  & $\checkmark$ & & &  $\checkmark$  \\ 
        ICML15 & Ioffe et al.~\cite{ioffe2015batch} &  & $\checkmark$  &  & &   & & & &  & $\checkmark$ & & &  $\checkmark$  \\  
        \hline
            \wugetable{CVPR22} & Wang et al.~\cite{wang2022semi} &  & $\checkmark$ & & & $\checkmark$ & &  & &  & $\checkmark$ & $\checkmark$ & $\checkmark$ & $\checkmark$ \\ 
       \wugetable{ACM MM21} & Chen et al.~\cite{chen2021transferrable}&  & $\checkmark$ & & & $\checkmark$ & &  & &  & $\checkmark$ & $\checkmark$ & $\checkmark$ & $\checkmark$   \\ 
       \wugetable{ACM MM21} & Luo et al.~\cite{luo2021coco} &  & $\checkmark$ & & & $\checkmark$ & &  & &  & $\checkmark$ & $\checkmark$ & $\checkmark$ & $\checkmark$  \\ 
       \wugetable{CVPR21} & Xie et al.~\cite{xie2021propagate} &  & $\checkmark$ & & & $\checkmark$ & &   & & & $\checkmark$ & $\checkmark$ & $\checkmark$ & $\checkmark$ \\ 
       ICCV21 & Caron et al.~\cite{caron2021emerging} &  & $\checkmark$ &  & & $\checkmark$ & &  & &  & $\checkmark$ & $\checkmark$ & & $\checkmark$   \\ 
        \wugetable{ICCV21} & Van et al.~\cite{van2021unsupervised} &  & $\checkmark$ & & & $\checkmark$ & &   & & & $\checkmark$ & $\checkmark$ & $\checkmark$ & $\checkmark$ \\ 
        \wugetable{ICCV21} & Ayush et al.~\cite{ayush2021geography} &  & $\checkmark$ &  & & $\checkmark$ & &  & &  & $\checkmark$ & $\checkmark$ &$\checkmark$ & $\checkmark$ \\ 
         \wugetable{ICCV21} & Xie et al.~\cite{xie2021detco} &  & $\checkmark$ &  & & $\checkmark$ & &  & &  & $\checkmark$ & $\checkmark$ &$\checkmark$ & $\checkmark$  \\ 
       \wugetable{ICLR21} & Li et al.~\cite{li2020prototypical} &  & $\checkmark$ &  & & $\checkmark$ & &  & &  & $\checkmark$ & $\checkmark$ &$\checkmark$ & $\checkmark$  \\ 
       \wugetable{Arxiv21} & El et al.~\cite{el2021training}  &  & $\checkmark$ & & & $\checkmark$ & &  & &  & $\checkmark$ & $\checkmark$ & $\checkmark$ & $\checkmark$   \\ 
        CVPR20 & He et al.~\cite{he2020momentum} &  & $\checkmark$ & & & $\checkmark$ & &  & &  & $\checkmark$ & $\checkmark$ & $\checkmark$ & $\checkmark$ \\ 
        \wugetable{NeurIPS20} & Chen et al.~\cite{chen2020big} &  & $\checkmark$ & & & $\checkmark$ & &   & & & $\checkmark$ & $\checkmark$ & $\checkmark$ & $\checkmark$ \\ 
        Arxiv20 & Chen et al.~\cite{chen2020improved} &  & $\checkmark$ & & & $\checkmark$ &  & &  & & $\checkmark$ & $\checkmark$ & $\checkmark$ & $\checkmark$ \\ 
        \hline
        
        \wugetable{ Arxiv23} & Zhang et al.~\cite{zhang2023robust} &  &  & & & $\checkmark$ &  & & &  & $\checkmark$ & $\checkmark$  & &  $\checkmark$  \\ 
        \wugetable{ AAAI22 } & Zhang et al.~\cite{zhang2022semi} &  &  & & & $\checkmark$ &  & & &  & $\checkmark$ & $\checkmark$  & &  $\checkmark$  \\ 
        \wugetable{ CVPR22} & Liu et al.~\cite{liu2022perturbed} &  &  & & & $\checkmark$ &  & & &  & $\checkmark$ & $\checkmark$  & &  $\checkmark$  \\ 
        \wugetable{ CVPR22} & Liu et al.~\cite{liu2022unbiased } &  &  & & & $\checkmark$ &  & & &  & $\checkmark$ & $\checkmark$  & &  $\checkmark$  \\ 
        \wugetable{ ECCV22} & Wang et al.~\cite{wang2022unsupervised} &  &  & & & $\checkmark$ &  & & &  & $\checkmark$ & $\checkmark$  & &  $\checkmark$  \\ 
        \wugetable{ICML22} & Wang et al.~\cite{wang2022self} &  &  & & & $\checkmark$ & &  & &  & $\checkmark$ & $\checkmark$ &  &  $\checkmark$   \\ 
        \wugetable{NeurIPS22} & Wang et al.~\cite{wang2022efficient} &  &  & & & $\checkmark$ & &  & &  & $\checkmark$ & $\checkmark$ & $\checkmark$ &    \\ 
        \wugetable{Arxiv22} & Wang et al.~\cite{wang2022learn} &  &  & & & $\checkmark$ & &  & &  & $\checkmark$ & $\checkmark$ & $\checkmark$ &   \\ 
        \wugetable{Arxiv22} & Tomasev et al.~\cite{tomasev2022pushing} &  &  & & & $\checkmark$ &  & &  & & $\checkmark$ & $\checkmark$ & & $\checkmark$ \\ 
        \wugetable{ Arxiv22} & Cai et al.~\cite{cai2022semi} &  &  & & & $\checkmark$ &  & & &  & $\checkmark$ & $\checkmark$  & &  $\checkmark$  \\ 
        \wugetable{ Arxiv22} & Kim et al.~\cite{kim2022pose} &  &  & & & $\checkmark$ &  & & &  & $\checkmark$ & $\checkmark$  & &  $\checkmark$  \\ 
        \wugetable{ CVPR21} & Cai et al.~\cite{cai2021exponential} &  &  & & & $\checkmark$ &  & & &  & $\checkmark$ & $\checkmark$  & &  $\checkmark$  \\ 
         \wugetable{ CVPR21} & Wang et al.~\cite{wang2021self} &  &  & & & $\checkmark$ &  & & &  & $\checkmark$ & $\checkmark$  & &  $\checkmark$  \\ 
         \wugetable{ CVPR21} & Yang et al.~\cite{yang2021interactive } &  &  & & & $\checkmark$ &  & & &  & $\checkmark$ & $\checkmark$  & &  $\checkmark$  \\ 
         \wugetable{ CVPR21} & Deng et al.~\cite{deng2021unbiased } &  &  & & & $\checkmark$ &  & & &  & $\checkmark$ & $\checkmark$  & &  $\checkmark$  \\ 
         &                                                    & $p$ & $\{x_i\}$ & $x_0$ & $y$ & $\{W_i\}$ & $\{\nabla_{x_i}\}$ & $\{\nabla_{W_i}\}$ & $l$     & Data & Model & Loss                       &\ \ DE\ \ & MA  \\\hline
        ICCV21 & Feng et al.~\cite{feng2021temporal} &  &  & & & $\checkmark$ & &   & & & $\checkmark$ & $\checkmark$ & $\checkmark$ & $\checkmark$ \\ 
        ICCV21 & Chen et al.~\cite{chen2021empirical} &  &  & & & $\checkmark$ &  & &  & & $\checkmark$ & $\checkmark$ & & $\checkmark$ \\ 
        \wugetable{ ICCV21} & Meng et al.~\cite{meng2021spatial} &  &  & & & $\checkmark$ &  & & &  & $\checkmark$ & $\checkmark$  & &  $\checkmark$  \\ 
         \wugetable{ ICCV21} & Wang et al.~\cite{wang2021multi} &  &  & & & $\checkmark$ &  & & &  & $\checkmark$ & $\checkmark$  & &  $\checkmark$  \\ 
         ICCV21 & Xu et al.~\cite{xu2021end} &  &  & & & $\checkmark$ &  & &  & & $\checkmark$ & $\checkmark$ & & $\checkmark$ \\ 
        \wugetable{ICLR21} & Mitrovic et al.~\cite{mitrovic2020representation} &  &  & & & $\checkmark$ &  & &  & & $\checkmark$ & $\checkmark$ & & $\checkmark$ \\ 
        \wugetable{Arxiv21} & Liu et al.~\cite{liu2021unbiased} &  &  & & & $\checkmark$ &  & &  & & $\checkmark$ & $\checkmark$ & & $\checkmark$ \\ 
        CVPR20 & Yuan et al.~\cite{yuan2020revisiting} &  &  & & & $\checkmark$ & & &  &  & $\checkmark$ & $\checkmark$ & $\checkmark$ & \\ 
        NeurIPS20 & Grill et al.~\cite{grill2020bootstrap} &  &  & & & $\checkmark$  &  & &  &  & $\checkmark$ & $\checkmark$ & & $\checkmark$  \\ 
        CVPR19 & Li et al.~\cite{li2019learning}  &  &  & & & $\checkmark$ &   & & & & $\checkmark$ & $\checkmark$ & $\checkmark$ & $\checkmark$ \\ 
        CVPR19 & Yang et al.~\cite{yang2019snapshot} &  &  & & & $\checkmark$ & &  & &  & $\checkmark$ & $\checkmark$ & $\checkmark$ &    \\ 
        \wugetable{CVPR19} &Cai et al.~\cite{cai2019exploring } &  &  & & & $\checkmark$ &  & &  & & $\checkmark$ & $\checkmark$ & & $\checkmark$ \\ 
         DLMIA18 & Perone et al.~\cite{perone2018deep} &  &  & & & $\checkmark$ &  & & &  & $\checkmark$ & $\checkmark$  & &  $\checkmark$  \\ 
        \wugetable{ ECCV18} & Shi et al.~\cite{shi2018transductive} &  &  & & & $\checkmark$ &  & & &  & $\checkmark$ & $\checkmark$  & &  $\checkmark$  \\ 
        ICML18 & Furlanello et al.~\cite{furlanello2018born} &  &  & & & $\checkmark$ & &  & &  & $\checkmark$ & $\checkmark$ & $\checkmark$ &   \\ 
        NeurIPS17 & Tarvainen et al.~\cite{tarvainen2017mean} &  &  & & & $\checkmark$ &  & & &  & $\checkmark$ & $\checkmark$  & &  $\checkmark$  \\ 
        
        \wugetable{CVPR20} & Poggi et al.~\cite{poggi2020uncertainty} &  &  & & & $\checkmark$ & &  & & & $\checkmark$ & & $\checkmark$ &    \\ 
        \wugetable{NeurIPS20} & Liu et al.~\cite{liu2020loss} &  &  & & & $\checkmark$ & &  & & & $\checkmark$ & & $\checkmark$ &    \\ 
        \wugetable{CVPR19} & Yu et al.~\cite{yu2019deep} &  &  & & & $\checkmark$ & &  & & & $\checkmark$ & & $\checkmark$ &    \\ 
         \wugetable{ECCV18} & Ilg et al.~\cite{ilg2018uncertainty} &  &  & & & $\checkmark$ & &  & & & $\checkmark$ & & $\checkmark$ &    \\ 
        ICLR17 & Huang et al.~\cite{huang2017snapshot} &  &  & & & $\checkmark$ & &  & & & $\checkmark$ & & $\checkmark$ &    \\ 
        \wugetable{Arxiv17} & Chen et al.~\cite{chen2017checkpoint} &  &  & & & $\checkmark$ & &  & & & $\checkmark$ & & $\checkmark$ &\\ 

        \wugetable{Arxiv22} & Guo et al.~\cite{guo2022stochastic} &  &  & & & $\checkmark$ & & &  &  & $\checkmark$ & & & $\checkmark$\\
        \wugetable{ICML21} & Von et al.~\cite{von2020neural} &  &  & & & $\checkmark$ & &   & & & $\checkmark$ & &  & $\checkmark$ \\ 
        \wugetable{NeurIPS21} & Cha et al.~\cite{cha2021swad} &  &  & & & $\checkmark$ & & &  &  & $\checkmark$ & & & $\checkmark$\\
        \hline
        Arxiv20 & Zhang et al.~\cite{zhang2020swa}  &  &  & & & $\checkmark$ &  & & &  & $\checkmark$ & &  & $\checkmark$ \\
        ICLR19 & Athiwaratkun et al.~\cite{athiwaratkun2018there} &  &  & & & $\checkmark$ & &  & &  & $\checkmark$ & & & $\checkmark$\\
        \wugetable{ICML19} & Yang et al.~\cite{yang2019swalp} &  &  & & & $\checkmark$ & & &  &  & $\checkmark$ & & & $\checkmark$\\
         NeurIPS19 & Maddox et al.~\cite{maddox2019simple} &  &  & & & $\checkmark$ & &  & &  & $\checkmark$ & & & $\checkmark$\\ 
         NeurIPS18 & Garipov et al.~\cite{garipov2018loss} &  &  & & &  $\checkmark$ & &  & & & $\checkmark$ & & & $\checkmark$\\
        \hline
		Arxiv22 & Li et al.~\cite{li2022equalized} &  &  &  & & &$\checkmark$ & & & &  & $\checkmark$ &  $\checkmark$ &   \\ 
         CVPR21 & Tan et al.~\cite{tan2021equalization} &  &  &  & & &$\checkmark$ & & & &  & $\checkmark$ & $\checkmark$  &   \\ 
        \wugetable{ECCV20} & Kwon et al.~\cite{kwon2020backpropagated} &  &  &  & & &$\checkmark$ & & & &  & $\checkmark$ &     &$\checkmark$  \\
        \hline
		\wugetable{ NeurIPS22} & Xie et al.~\cite{xie2022adan} &  &  &  & & & &  $\checkmark$ & & & $\checkmark$ &  &   & $\checkmark$  \\ 
		\wugetable{ ICLR21} & Heo et al.~\cite{heo2020adamp} &  &  &  & & & &  $\checkmark$ & & & $\checkmark$ &  &   & $\checkmark$  \\ 
        ICLR20 & Liu et al.~\cite{liu2019variance} &  &  &  & & & &  $\checkmark$ & & & $\checkmark$ &  &   & $\checkmark$  \\ 
       \wugetable{ IJCAI20} & Chen et al.~\cite{chen2018closing} &  &  &  & & & &  $\checkmark$ & & & $\checkmark$ &  &   & $\checkmark$  \\ 
       ICLR19 & Loshchilov et al.~\cite{loshchilov2017decoupled} &  &  & & &  &  & $\checkmark$ & & & $\checkmark$ &  &   & $\checkmark$  \\ 
       Arxiv19 & Ding et al.~\cite{ding2019adaptive} &  &  & & &  &  & $\checkmark$ & & & $\checkmark$ &  &   & $\checkmark$  \\ 
        \wugetable{ICLR16} & Dozat et al.~\cite{dozat2016incorporating} &  &  & & &  &  & $\checkmark$ & & & $\checkmark$ &  &   & $\checkmark$ \\ 
        ICLR15 & Kingma et al.~\cite{kingma2014adam}  &  &  & & &  &   & $\checkmark$ & & &$\checkmark$ &   & & $\checkmark$ \\ 
        ICML13 & Sutskever et al.~\cite{sutskever2013importance} &  &  & & & &   & $\checkmark$&  & &$\checkmark$ &   & & $\checkmark$ \\ 
        JMLR11 & Duchi et al.~\cite{duchi2011adaptive} &  &  & & &  &  & $\checkmark$ & & &$\checkmark$ &  &   & $\checkmark$\\ 
        \hline
        \wugetable{AAAI22} & Jiang et al.~\cite{jiang2021delving} &  &  & & &  &  &  & $\checkmark$& $\checkmark$& &  $\checkmark$ &  $\checkmark$&  \\
        \wugetable{ Arxiv22} & Yue et al.~\cite{yue2022ctrl}  &  &  & & &  &  &  & $\checkmark$& $\checkmark$& &  $\checkmark$ &  $\checkmark$&  $\checkmark$  \\
        \wugetable{ECCV20} & Chu et al.~\cite{chu2020neural}  &  &  & & &  &  &  & $\checkmark$& $\checkmark$& &  $\checkmark$ &  $\checkmark$& \\
         \wugetable{ECML21} &Xu et al.~\cite{xu2021small} &  &  & & &  &  &  & $\checkmark$& $\checkmark$& &  $\checkmark$ &  $\checkmark$& $\checkmark$  \\ 
         \wugetable{ICCV21} & Lazarou et al.~\cite{lazarou2021iterative}&  &  & & &  &  &  & $\checkmark$& $\checkmark$& &  $\checkmark$ &  $\checkmark$&   $\checkmark$ \\
         \wugetable{Arxiv21} & Xia et al.~\cite{xia2021sample}&  &  & & &  &  &  & $\checkmark$& $\checkmark$& &  $\checkmark$ &  $\checkmark$& $\checkmark$   \\
         \wugetable{CVPR19} & Huang et al.~\cite{huang2019o2u} &  &  & & &  &  &  & $\checkmark$& $\checkmark$& &  $\checkmark$ &  $\checkmark$&  $\checkmark$  \\ 
        \hline
        NeurIPS22 & Yang et al.~\cite{yang2022recursivemix} &  $\checkmark$  &  &  $\checkmark$ & $\checkmark$ &  & &  & & $\checkmark$  &  & $\checkmark$ & $\checkmark$ &   \\
        \hline
\end{xtabular}
\end{strip}

\begin{figure*}[t]
	\vspace{0pt}
	\begin{center}
		\setlength{\fboxrule}{0pt}
		\fbox{\includegraphics[width=1\textwidth]{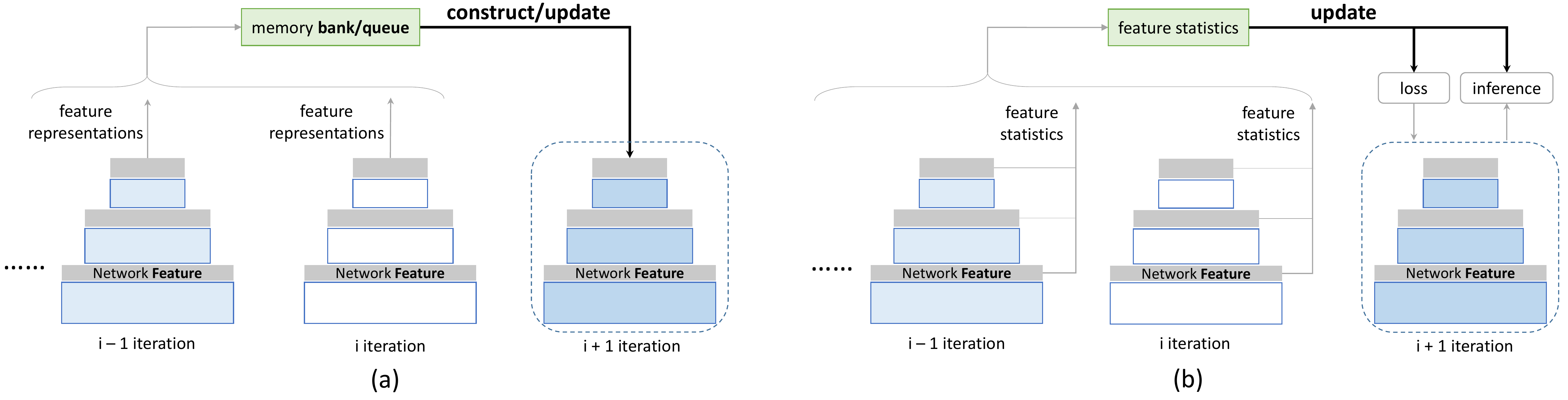}}
	\end{center}	
	\vspace{-15pt}
	\caption{ \textbf{Two streams of employing historical intermediate feature representation.} (a) Record the historical instance-level feature representations by memory bank or queue to construct/update sufficient and critical training pairs. (b) Memorize feature statistics which are further utilized in training loss or inference.
	}
	\label{fig_intermediate_feature_representation}
\end{figure*}

\noindent\textbf{Intermediate Feature Representation.} There are mainly two streams  and the specific structure comparison is shown in Fig.~\ref{fig_intermediate_feature_representation}.

{
\ylfChecked{(1) One stream is to record the instance-level feature representations aiming at constructing sufficient and critical training pairs}
(see Fig.~\ref{fig_intermediate_feature_representation} (a)). 
The common practices of obtaining feature representations via computing the current mini-batch lack diversity and efficiency. 
Sprechmann et al.~\cite{sprechmann2018memory} introduce an episodic memory module to store examples and modify network parameters based on the current input context. Wang et al.~\cite{wang2020cross} extend this concept with the introduction of a cross-batch memory (XBM) that records the instance embeddings of past iterations, allowing the algorithm to gather more hard negative pairs. 
Influenced by XBM, some work has embedded the memory bank into the network to adapt various tasks~\cite{deng2022insclr, yang2022online, deng2021variational, li2020unsupervised,feng2021exploring, liu2022memory,xiao2017joint,alonso2021semi,lee2021weakly,li2022recurrent,liu2021one,yang2018learning, yang2019visual,el2021training,liu2022densely,wang2022contrastive,wen2021false,liu2022learning,liu2021noise,ortego2021multi,jin2021mining,yang2022cross,zhong2019invariance}. To enhance the diversity of semantic differences, Liu et al.~\cite{liu2022densely} construct a memory bank using historical intra-class embedding representations. Motivated by the issue of slow feature drift, Wang et al.~\cite{wang2022contrastive} propose the quantization code memory bank to lower feature drift to use historical feature representation effectively. Wen et al.~\cite{wen2021false} offer a delayed update memory bank that ranks instances in the current mini-batch for machine learning multi-partite ranking tasks. Liu et al.~\cite{liu2022learning} also employ slow feature drift to accumulate the historical modality-specific proxies (MS-Proxies) stored by the memory bank, thereby enhancing the diversity of these MS-Proxies. 
\ylfChecked{Methods have also been proposed to address the issue of label noise that arises from storing instances or features in memory banks~\cite{liu2021noise,ortego2021multi}.}
\ylfChecked{Liu et al.~\cite{liu2021noise} take a different approach by calculating the probability of having clean labels and choosing high-scoring ones to extract features into the memory bank, resulting in producing superior historical representation and increasing resistance to noise.}
Some works skillfully combine the memory banks with specific downstream tasks, which promote the performance of the model. 
\ylfChecked{
In the field of semantic image segmentation, Jin et al.~\cite{jin2021mining} use a memory bank to store different class representations. Yang et al.~\cite{yang2022cross} employ a pixel queue and a region queue to construct the memory bank for modeling pixel dependencies and feature relations between pixels and class region, respectively. 
The work enhances global semantic relations through historical feature representation.}
In domain adaptation, Zhong et al.~\cite{zhong2019invariance} use a memory bank to store the feature maps of unlabeled data from the target domain and apply an exemplar-invariance objective from the source domain to bridge the gap.}

{Different from storing some instance representations~\cite{wang2020cross}, Wu et al.~\cite{wu2018improving,wu2018unsupervised} develop the memory bank mechanism to memorize the representations for all images to perform Neighborhood Component Analysis (NCA)~\cite{wu2018improving} or Noise-Contrastive Estimation (NCE)~\cite{wu2018unsupervised} optimization, instead of exhaustively computing these embeddings every time. Following~\cite{ wu2018unsupervised}, 
some works~\cite{ misra2020self, tian2020contrastive, yin2021instance, dai2021dual,wang2022semi,ko2021learning} also employ the memory bank to store all image or class representations. Yin et al.~\cite{yin2021instance} propose Class Feature Banks (CFB), which use historical feature representation to record and update each class's diverse representation. Notably, previous works on memory bank~\cite{wu2018improving,wu2018unsupervised} may not be suitable for online feature learning, as the stored features in the memory bank cannot be updated through the back-propagation of gradients. To address this, Dai et al.~\cite{dai2021dual} propose an immediate memory bank that stores instant features of all the samples and updates them regularly. However, these methods~\cite{wang2020cross,wen2021false} primarily focus on categories that have already occurred and have a weaker impact on categories that have not yet appeared during testing. To address this issue, Ko et al.~\cite{ko2021learning} propose storing both embedding features and category weights in the memory bank to reduce the emphasis on seen classes. 
}

It is noteworthy that these representations~\cite{wang2020cross} will not be updated with momentum encoder before inputting into the memory bank. In contrast, He et al.~\cite{he2020momentum} and Chen et al.~\cite{chen2020improved} introduce the concept of the momentum memory bank, a queue of data samples in which features are progressively replaced and updated by momentum. Some subsequent work~\cite{li2021spatial, el2021training ,chen2021transferrable , zhou2022regional, luo2021coco,chen2020big,van2021unsupervised,xie2021propagate,li2020prototypical,ayush2021geography, xie2021detco,zheng2021group,zhuang2019local} has built upon these ideas. Li et al.~\cite{li2021spatial} use the momentum memory bank to design a novel spatial assembly network, which can learn features of different spatial variations.
Following~\cite{wang2020cross, he2020momentum}, El et al.~\cite{el2021training} use an offline memory bank and a momentum encoder to improve the performance of image retrieval baselines. In addition, Zheng et al.~\cite{zheng2021group} introduce a weighted contrastive loss to calculate the similarity of characteristics between the mini-batch and memory bank, which speeds up the optimization of the model. 
In the context of domain adaptation, Chen et al.~\cite{chen2021transferrable}  address the problem of domain adaptation by using source memory and target memory to record the class characteristics of the source domain and the pseudo-label characteristics of the target domain, respectively.

(2)  {
\ylfChecked{Another stream is to memorize statistical-level feature representations for utilizing in training loss or inference}
~(see Fig.~\ref{fig_intermediate_feature_representation} (b)). 
The center loss~\cite{wen2016discriminative} and its different variants~\cite{he2018triplet,li2019angular,min2020adversarial,zhao2020deep,narayan20193c,chen2021free,li2021frequency, keshari2020generalized, jing2021cross,fan2018associating,liu2020leveraging} attempt to record and update the category representations related to history. The goal is to minimize the intra-class distance and maximize the inter-class distance through the historical category representation. Center loss~\cite{wen2016discriminative} updates the category-level feature centers and penalizes the distances between the deep features and their corresponding class centers, promoting stronger intra-class compactness. The specific formula is as follows:
\begin{equation}
\begin{aligned}
\mathcal{L}_{\text {center }}=\frac{1}{2} \sum_{i=1}^{b}\left\|x_{i}-c_{y_{i}}\right\|_{2}^{2}.
\end{aligned}
\end{equation}
The center loss, represented by~$\mathcal{L}_{center}$, is calculated over a mini-batch of size $b$. The calculation involves subtracting the class center~$c_{y_{i}}$ from the~$i$-th feature vector~$x_{{i}}$, and~$y_{i}$ represents class index. The intra-class distance is then computed as~$\left\|x_{i}-c_{y_{i}}\right\|_{2}^{2}$. On each iteration, the class center~$c_{y_{i}}$ is updated, condensing the class characteristics over different historical periods. To further balance the distance between intra-class and inter-class, He et al.~\cite{he2018triplet} combine the advantage of triplet loss and center loss, which expressly seeks to enhance both inter-class separability and intra-class compactness. Li et al.~\cite{li2019angular} use angular distance instead of Euclidean distance used by~\cite{he2018triplet}, which gives more distinct discriminating limitations in the angle space. Later, Min et al.~\cite{min2020adversarial} combine the adversarial approach ACoL~\cite{ zhang2018adversarial } with center loss, which utilizes the majority of historical features statistics and also avoids the domination by some historical features statistics. Motivated by~\cite{wen2016discriminative} and~\cite{he2018triplet}, Zhao et al.~\cite{zhao2020deep} propose Hard Mining Center-Triplet Loss (HCTL). HCTL chooses the hardest positive/negative samples on whether it has the same label as each center, which uses the historical feature statistics to perform hard sample mining. Later, Li et al.~\cite{li2021frequency} selectively compress intra-class features into various embedding spaces, attempting to handle separately historical feature statistics for better optimization.
However, the center loss is designed for a single-label sample, making it difficult to extend to weakly-supervised multi-label tasks. Narayan et al.~\cite{narayan20193c} address this problem using attention-based per-class feature aggregation, calculating an attention center loss that employs more historical information and effectively reduces the intra-class distance of multi-label. 
Besides,~\cite{keshari2020generalized} and~\cite{chen2021free} expand center loss to zero-shot learning. Notably, ~\cite{chen2021free} introduce the self-adaptive margin center (SAMC) loss to learn the historical fine-grained and visual-semantic embedding class centers. 
Jing et al.~\cite{jing2021cross} show that it is also possible to learn class centers from cross-modal samples. Cross-modal characteristics are transferred into the same representation space, then calculate the center loss, which employs the historical cross-modal feature to reduce the gaps.}

{Aside from statistics for classes, there are other ways to utilize historical statistics for features, e.g., moving mean and variance. Caron et al.~\cite{caron2021emerging} maintain the first-order batch statistics by adding a bias term to the teacher network, successfully avoiding the training collapse. {According to Ioffe et al.~\cite{ioffe2015batch}, during model inference, the mean and variance statistics computed and accumulated inside each Batch Normalization (BN) layer across all levels of features are utilized. The various variations of~\cite{ioffe2015batch, ioffe2017batch, li2016revisiting, salimans2016weight,arpit2016normalization} can lead to faster convergence and improved generalization of the model.}}

\begin{figure*}[t]
	\vspace{0pt}
	\begin{center}
		\setlength{\fboxrule}{0pt}
		\fbox{\includegraphics[width=1\textwidth]{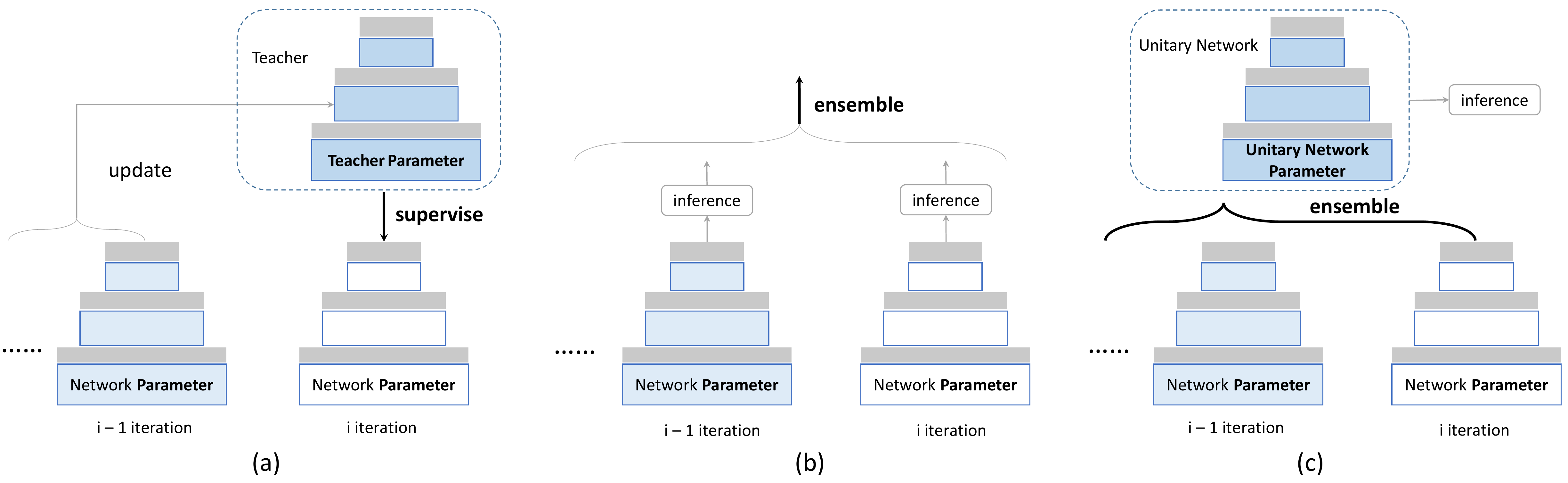}}
	\end{center}	
	\vspace{-15pt}
	\caption{ \textbf{Three types of using historical model parameters.} (a) Construct teachers based on historical student parameters to supervise the current student effectively. (b) Utilize ensemble results by inferring from several historical models. (c) Create a unitary ensemble architecture using the historical model parameters for efficient inference. 
	}
	\label{fig_hl_model_parameters}
\end{figure*}

\noindent\textbf{Model Parameter.} The historical usage of model parameters can be divided into three major groups, and a comparison of these three structures is shown in Fig.~\ref{fig_hl_model_parameters}.

(1) Constructing the teachers from past models  {to supervise the students effectively (see Fig.~\ref{fig_hl_model_parameters} (a)).} The popular practice is to build a teacher network by leveraging the historical student parameters in unsupervised~\cite{caron2021emerging,he2020momentum,chen2021empirical,feng2021temporal,grill2020bootstrap,chen2020improved, el2021training ,chen2021transferrable ,luo2021coco,chen2020big,van2021unsupervised,xie2021propagate,li2020prototypical,ayush2021geography,mitrovic2020representation,cai2019exploring, deng2021unbiased, xie2021detco,tomasev2022pushing}, supervised~\cite{yuan2020revisiting,li2019learning,yang2019snapshot,furlanello2018born,wang2022efficient,wang2022learn,wang2022self} and semi-supervised~\cite{tarvainen2017mean,cai2022semi,wang2022unsupervised,shi2018transductive,wang2021multi, kim2022pose, yu2019uncertainty, meng2021spatial, wang2021self ,wang2022semi,liu2022perturbed, cui2019semi, perone2018deep,zhang2022semi, xu2021end, yang2021interactive, liu2021unbiased, liu2022unbiased, cai2021exponential, zhang2023robust} learning. In the unsupervised learning, previous works~\cite{he2020momentum,chen2020improved} have shown the effectiveness of unsupervised learning exploiting historical parameters. 
Chen et al.~\cite{chen2021empirical} focus on building a self-supervised framework using ViTs and achieve good results. Feng et al.~\cite{feng2021temporal} propose a method called Temporal Knowledge Consistency (TKC), which integrates instance temporal consistency into the teacher model and dynamically combines the knowledge from temporal teacher models to select valuable information. Grill et al.~\cite{grill2020bootstrap} introduce a self-supervised approach called Bootstrap Your Own Latent (BYOL), in which the representation of the target network is predicted using a trained online network. Caron et al.~\cite{caron2021emerging} find that self-supervised ViT features provide more explicit information about tasks such as semantic segmentation, which employ a momentum encoder to predict the teacher network's output. 
{In the realm of supervised learning, researchers have proposed various approaches for employing historical parameters. Yuan et al.~\cite{yuan2020revisiting} suggest allowing the student model to learn from itself or a predefined regularization distribution, without relying on a teacher model. 
Li et al.~\cite{li2019learning} train models with synthetic noisy labels to optimize the model through the generated noise. Yang et al.~\cite{yang2019snapshot} use the last snapshot of each cycle as the teacher for the next cycle, in order to capture consistent information. Furlanello et al.~\cite{furlanello2018born} propose to improve performance by training multiple identical student models using different random seeds and the supervision of the previous generation. Wang et al.~\cite{wang2022efficient,wang2022learn} enhance distillation performance and reduce training costs by training the student model with historical snapshots from the same training trajectory, and by incorporating a self-attention module to adapt the weights of the snapshots. Another work~\cite{wang2022self} uses historical parameters as the teacher to update the current model through EMA during adversarial training.}

{In the context of semi-supervised learning, Tarvainen et al.~\cite{tarvainen2017mean} propose Mean teacher, which updates the teacher model based on the EMA of the historical student model’s parameters during training. Later, several  work uses~\cite{ cai2022semi, wang2022unsupervised, shi2018transductive,wang2021multi, kim2022pose, yu2019uncertainty, meng2021spatial, wang2021self ,wang2022semi,liu2022perturbed, cui2019semi, perone2018deep,zhang2022semi, xu2021end, yang2021interactive , liu2021unbiased, liu2022unbiased} or improves~\cite{ cai2021exponential, zhang2023robust} on the Mean teacher.
Cai et al.~\cite{cai2021exponential} propose exponential moving average normalization (EMAN) to replace the BN of the teacher model. EMAN aims to reduce the cross-sample dependency and model parameter mismatch to utilize historical student parameters efficiently. Then, Cai et al.~\cite{cai2022semi} extend the Mean teacher to the pure ViT for semi-supervised learning, having more scalability when using historical parameters and reducing the performance gap with fully supervised. 
Notably, Zhang et al.~\cite{zhang2023robust} propose the adaptive exponential moving average (AEMA) by adding an adjustment parameter~$\delta$ on the basis of EMA as follows:
\begin{equation}
\begin{aligned}
W_{i}^{t}  =\alpha\delta W_{i}^{t-1}+(1-\alpha\delta) W_{i}^{t}.
\end{aligned}
\end{equation}
The model parameters of~$t$-th iteration~$W_{i}^{t}$ are updated through the combination of historical model parameters~$W_{i}^{t-1}$ by momentum term~$ \alpha$. In general, AEMA generates better quality pseudo labels by adaptively adjusting historical and current model parameters.}
The teacher models can provide informative supervision to the original model, leading to significant improvement in the performance of student models.

{(2) {Directly exploiting ensemble results in inference from multiple past models (see Fig.~\ref{fig_hl_model_parameters} (b)).}
Huang et al.~\cite{huang2017snapshot} and Chen et al.~\cite{chen2017checkpoint} use a set of model snapshots (or checkpoints) to perform multiple evaluations for each test sample and ensemble the predictions as the final result. 
Later, some works extend~\cite{huang2017snapshot} to various tasks or fields: short-term load forecasting~\cite{ chen2018short }, uncertainty estimation ~\cite{ilg2018uncertainty, poggi2020uncertainty}, MR images segmentation~\cite{ wolterink2018automatic}, image denoising~\cite{yu2019deep},  adversarial training~\cite{liu2020loss}, object detection~\cite{ xu2020multi }, and medical image registration~\cite{gong2022uncertainty}. 
It is noteworthy that Chen et al.~\cite{ chen2018short } expand the~\cite{huang2017snapshot} to two stages, which involves integrating  models with different initial parameters. After obtaining all the snapshot models, the outputs of the models are averaged and produce the final forecast, using historical predictions from distinct training periods to reduce the standard deviation and improve generalization capability.}

(3) \ylfChecked{Building unitary ensemble architecture from previous models in inference}
(see Fig.~\ref{fig_hl_model_parameters} (c)). {Different from~\cite{huang2017snapshot,chen2017checkpoint} where multiple networks are adopted for inference,~\cite{zhang2020swa,athiwaratkun2018there,izmailov2018averaging,yang2019swalp,cha2021swad,guo2022stochastic,garipov2018loss,maddox2019simple,von2020neural} attempt to average/ensemble the historical model parameters directly and obtain a final single structure for efficient deployment.} 
{For instance, Zhang et al.\cite{zhang2020swa} use cyclical learning rates to train extra epochs and average them to improve performance. Athiwaratkun et al.\cite{athiwaratkun2018there} try to identify high-accuracy pathways between loss functions to find high-performing ensembles in a short time. Izmailov et al.\cite{izmailov2018averaging} propose Stochastic Weight Averaging (SWA), which follows the trajectory of SGD to average multiple points and achieve better generalization performance. To adapt SWA to low precision training, Yang et al.~\cite{yang2019swalp} propose stochastic weight averaging in low-precision training (SWALP) to quantize gradient accumulators and velocity vectors during training. 
\ylfChecked{In another direction, Cha et al.~\cite{cha2021swad} extend SWA to Domain Generalization and aim to solve the problem of inaccurate approximation of flat minima that results from a small number of SWA sampling weights in high-dimensional parameter space.}
\ylfChecked{Further, Guo et al.~\cite{guo2022stochastic} propose a method that utilizes a series of weight averaging (WA) operations to reveal global geometric structures, leading to improved optimization compared to SGD.}
Garipov et al.~\cite{garipov2018loss} use averaging to accelerate convergence within each cycle of a cyclical learning rate schedule. Maddox et al.~\cite{maddox2019simple} average Bayesian models that are sampled from a Gaussian distribution. \ylfChecked{Von et al.~\cite{von2020neural} optimize independently and average copies of a subset of neural network parameters, such as weights in the late processing of training, to improve generalization subsequently.}}
\begin{table*}[htp]
	\vspace{0pt}
	\renewcommand\arraystretch{1.2}
	\centering
	\footnotesize \setlength{\tabcolsep}{4.pt}
    \caption{Related works of optimizers that rely on the statistics of historical gradients of the model parameters in a way of EMA for speeding up the training convergence. $W^{t-1}$ is not especially considered as any optimization step for neural networks involves updating the old network parameter.}
	\resizebox{0.9\textwidth}{!}
	{
        \begin{tabular}{l|l|l|l}
        \hline
        \multirow{1}{*}{Publication}& \multirow{1}{*}{Method}& \multirow{1}{*}{Update $W^{t}$} & \multirow{1}{*}{Historical Variable} \\ 

        \hline
       JMLR2011&Adagrad & $\begin{array}{l}
v^{t}=\sum_{0}^{t} (\nabla_{W_{i}^{t}})^2\\
W_{i}^{t}=W_{i}^{t-1}-\alpha \nabla_{W_{i}^{t}} / \sqrt{v^{t}+\epsilon}\\
\end{array}$  &${v^{t}}$ \\ 
        \hline
        ICML2013 &SGD+Momentum & $\begin{array}{l}
        v^{t}=\beta_{2} v^{t-1}+\left(1-\beta_{2}\right) \nabla_{W_{i}^{t}} \\
        W_{i}^{t} =  W_{i}^{t-1}-\alpha^{t} v^{t} 
\end{array}$&${v^{t}}$  \\
        \hline 
        ICLR2015&Adam &$\begin{array}{l}
m^{t}=\beta_{1} m^{t-1}+\left(1-\beta_{1}\right) \nabla_{W_{i}^{t}} \\
v^{t}=\beta_{2} v^{t-1}+\left(1-\beta_{2}\right) (\nabla_{W_{i}^{t}})^{2} \\
\hat{m}^{t}=m^{t} /\left(1-\beta_{1}^{t}\right) \\
\hat{v}^{t}=v^{t} /\left(1-\beta_{2}^{t}\right) \\
W_{i}^{t} =  W_{i}^{t-1}-\hat{m}^{t}\alpha^{t} /\left(\sqrt{v^{t}}+\epsilon\right)
\end{array}$ &${m^{t}}$,~$v^{t}$  \\
        \hline 
        ICLR2016 &Nadam & $\begin{array}{l}
m^{t}=\beta_{1} m^{t-1}+\left(1-\beta_{1}\right) \nabla_{W_{i}^{t}} \\
v^{t}=\beta_{2} v^{t-1}+\left(1-\beta_{2}\right) (\nabla_{W_{i}^{t}})^{2} \\
\hat{m}^{t} =\left(\beta_{1}^{t+1} {m}^{t} /\left(1-\prod_{j=1}^{t+1} \beta_{1}^{j}\right)\right)+\left(\left(1-\beta_{1}^{t}\right) {g}^{t} /\left(1-\prod_{j=1}^{t} \beta_{1}^{j}\right)\right) \\
\hat{{v}^{t}} = \beta_{2}{v}^{t} /\left(1-\beta_{2}^{t}\right) \\
W_{i}^{t}=W_{i}^{t-1}-\hat{m}^{t}\alpha^{t} /\left(\sqrt{v^{t}}+\epsilon\right)
\end{array}$&${m^{t}}$,~$v^{t}$ \\
        \hline 
        ICLR2019&AdamW &$\begin{array}{l}
m^{t}=\beta_{1} m^{t-1}+\left(1-\beta_{1}\right) \nabla_{W_{i}^{t}} \\
v^{t}=\beta_{2} v^{t-1}+\left(1-\beta_{2}\right) (\nabla_{W_{i}^{t}})^{2} \\
\hat{m}^{t}=m^{t} /\left(1-\beta_{1}^{t}\right) \\
\hat{v}^{t}=v^{t} /\left(1-\beta_{2}^{t}\right) \\
\eta^{t} = \text { SetScheduleMultiplier }(t)\\
W_{i}^{t}=W_{i}^{t-1}-\eta^{t}(\hat{m}^{t}\alpha^{t} /\left(\sqrt{v^{t}}+\epsilon\right)+\lambda {\theta}^{t-1})
\end{array}$ &${m^{t}}$,~$v^{t}$  \\
        \hline 
       Arxiv2019& AdaMod& $\begin{array}{l}
m^{t}=\beta_{1} m^{t-1}+\left(1-\beta_{1}\right) \nabla_{W_{i}^{t}} \\
v^{t}=\beta_{2} v^{t-1}+\left(1-\beta_{2}\right) (\nabla_{W_{i}^{t}})^{2} \\
\hat{m}^{t}=m^{t} /\left(1-\beta_{1}^{t}\right) \\
\hat{v}^{t}=v^{t} /\left(1-\beta_{2}^{t}\right) \\
s^{t}=\beta_{3} s^{t-1}+\left(1-\beta_{3}\right) \alpha^{t} /(\sqrt{\hat{v}^{t}}+\epsilon)\\
W_{i}^{t}=W_{i}^{t-1}-\hat{m}^{t}\min \left(\alpha^{t} /\left(\sqrt{\hat{v}^{t}}+\epsilon\right), s^{t} \right)\\
\end{array}
$&${m^{t}}$,~$v^{t}$,~$s^{t}$\\
        \hline 
        IJCAI2020&Padam &$\begin{array}{l}
m^{t}=\beta_{1} m^{t-1}+\left(1-\beta_{1}\right) \nabla_{W_{i}^{t}} \\
v^{t}=\beta_{2} v^{t-1}+\left(1-\beta_{2}\right) (\nabla_{W_{i}^{t}})^{2} \\
\hat{v}^{t}=\max(\hat{v}^{t-1} ,{v}^{t})\\
W_{i}^{t}=W_{i}^{t-1}-{m}^{t}\alpha^{t} /\hat{v}^{t}_{p}\quad p \in(0,1 / 2]\\
\end{array}
$&${m^{t}}$,~$v^{t}$  \\
        \hline 
        ICLR2020&RAdam & $\begin{array}{l}
m^{t}=\beta_{1} m^{t-1}+\left(1-\beta_{1}\right) \nabla_{W_{i}^{t}} \\
v^{t}=\beta_{2} v^{t-1}+\left(1-\beta_{2}\right) (\nabla_{W_{i}^{t}})^{2} \\
\hat{m}^{t}=m^{t} /\left(1-\beta_{1}^{t}\right) \\
\rho^{\infty} =2 /\left(1-\beta_{2}\right)-1\\
\rho^{t} = \rho^{\infty}-2 t \beta_{2}^{t} /\left(1-\beta_{2}^{t}\right)\\
W_{i}^{t}=\left\{\begin{array}{ll}
l^{t} =\sqrt{\left(1-\beta_{2}^{t}\right) / v^{t}} \\
r^{t}=\sqrt{\frac{\left(\rho^{t}-4\right)\left(\rho^{t}-2\right) \rho^{\infty}}{\left(\rho^{\infty}-4\right)\left(\rho^{\infty}-2\right) \rho^{t}}} & \left(\rho^{t}>4\right) \\
W_{i}^{t-1}-\alpha^{t} r^{t} \hat{m}^{t} {l}^{t} & \\
\\
W_{i}^{t-1}-\alpha^{t} \hat{m}^{t} & \left(\rho^{t} \leq 4\right)
\end{array}\right.
\end{array}$&${m^{t}}$,~$v^{t}$  \\        
\hline 
        ICLR2021&Adamp & $\begin{array}{l}
m^{t}=\beta_{1} m^{t-1}+\left(1-\beta_{1}\right) \nabla_{W_{i}^{t}} \\
v^{t}=\beta_{2} v^{t-1}+\left(1-\beta_{2}\right) (\nabla_{W_{i}^{t}})^{2} \\
{p}^{t}= {m}^{t} /\left(\sqrt{{v}^{t}}+\varepsilon\right)\\
{q}^{t}=\left\{\begin{array}{ll}
\Pi_{W_{i}^{t-1}}\left({p}^{t}\right) & \text { if } \cos \left(W_{i}^{t-1}, \nabla_{W_{i}^{t}}\right)<\delta / \sqrt{\operatorname{dim}({W})} \\
{p}_{t} & \text { otherwise }
\end{array}\right.\\
W_{i}^{t}=W_{i}^{t-1}-\alpha {q}^{t}
\end{array}$&${m^{t}}$,~$v^{t}$\\
        \hline 
        NeurIPS22&Adan & $\begin{array}{l}
m^{t}=\beta_{1} m^{t-1}+\left(1-\beta_{1}\right) \nabla_{W_{i}^{t}} \\
v^{t}=\beta_{2} v^{t-1}+\left(1-\beta_{2}\right) (\nabla_{W_{i}^{t}}-\nabla_{W_{i}^{t-1}}) \\
n^{t}=\left(1-\beta_{3}\right) n^{t-1}+\beta_{3}\left[\nabla_{W_{i}^{t}}+\left(1-\beta_{2}\right)\left(\nabla_{W_{i}^{t}}-\nabla_{W_{i}^{t-1}}\right)\right]^{2} \\
\alpha^{t}=\alpha /\left(\sqrt{n^{t}+\varepsilon}\right) \\
W_{i}^{t}=\left(1+\lambda_{t} \eta\right)^{-1}\left[W_{i}^{t-1}-\alpha^{t}\left(m^{t}+\left(1-\beta_{2}\right) v^{t}\right)\right]
\end{array}$&${m^{t}}$,~$v^{t}$,~$n^{t}$  \\
        \hline 
		\end{tabular}
	}
	\vspace{-35pt}
	\label{tab_gradients_summary}
\end{table*}

\begin{figure*}[t]
	\vspace{0pt}
	\begin{center}
		\setlength{\fboxrule}{0pt}
\fbox{\includegraphics[width=1\textwidth]{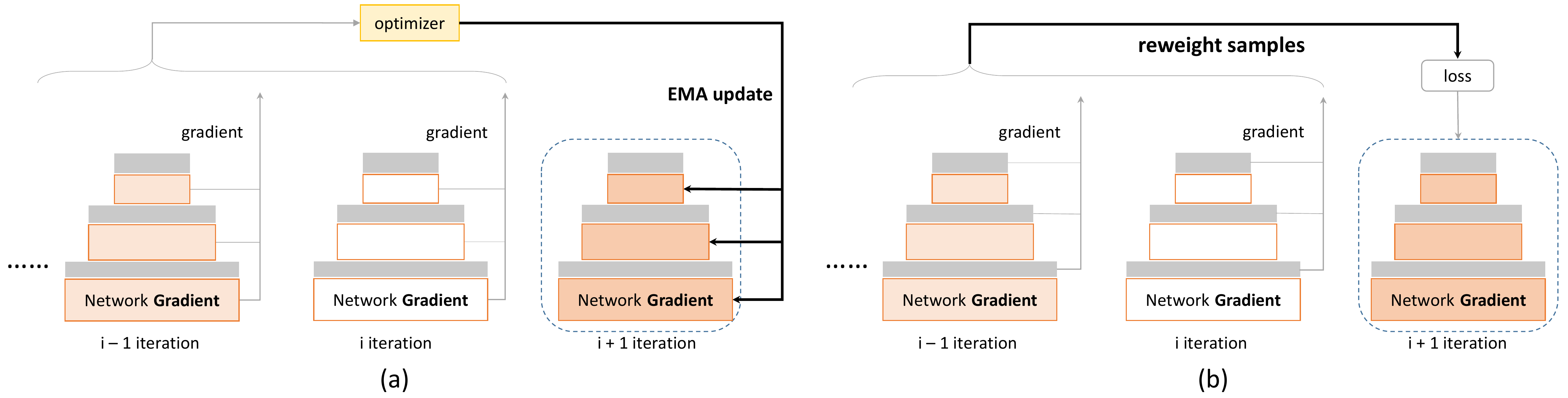}}
	\end{center}	
	\vspace{-13pt}
	\caption{ \textbf{Two ways to use historical gradient}. (a) Optimizers update model parameters by EMA with historical gradients of the model parameters. (b) The gradient-guided mechanism reweights samples using the historical gradients of different level attributes.}
	\label{fig_hl_gradient}
	\vspace{-0pt}
\end{figure*}

\noindent\textbf{Gradient.} 
\ylfChecked{There are two approaches within this direction--operating the gradients of the model parameters or gradients of all-level features.}
The comparison of the two structures is illustrated in Fig.~\ref{fig_hl_gradient}.

(1) The advanced optimizers commonly rely on the statistics of historical gradients of the model parameters in a way of EMA for speeding up the training convergence (see Fig.~\ref{fig_hl_gradient} (a)), where momentum SGD~\cite{sutskever2013importance} calculates the first-order statistics, Adagrad~\cite{duchi2011adaptive} utilizes the second-order one. The specific information, formulas, and historical variables involved in different optimizers are shown in Table~\ref{tab_gradients_summary}. 
Adam~\cite{kingma2014adam} and its variations are widely used for optimizing stochastic objective functions. They are effective algorithms that make use of both the mean and variance of the gradients. NAdam~\cite{dozat2016incorporating} incorporates Nesterov's accelerated gradient (NAG) technique to improve the quality of the learned model. AdamW~\cite{loshchilov2017decoupled} adds decoupled weight decay to improve generalization performance. 
AdaMod~\cite{ ding2019adaptive} uses adaptive upper bounds to limit the learning rates. 
Padam~\cite{chen2018closing} proposes the use of partial adaptive parameters to speed up model convergence. RAdam~\cite{liu2019variance} tries to correct the adaptive learning rate to maintain a constant variance. Adamp~\cite{heo2020adamp} modifies the practical step sizes to prevent the weight standard from increasing, and Adan~\cite{xie2022adan} introduces a Nesterov Momentum Estimation (NME) method to reduce training cost and improve performance.

(2) {The gradient-guided mechanism utilizes the historical gradients of various level features to reweight samples~(see Fig.~\ref{fig_hl_gradient} (b)). This method mainly focuses on dealing with long-tailed categories~\cite{tan2021equalization,li2022equalized} or abnormal samples~\cite{ kwon2020backpropagated} by using all levels of features. In order to reduce the accuracy gap between decoupled and end-to-end training techniques, Tan et al.~\cite{tan2021equalization} propose Equalization Loss v2 (EQL v2). EQL v2 rebalances the training process for each category independently and equally, which can be written as:
\begin{equation}
\begin{aligned}
{g^{j}}=\frac{\sum_{t=0}^{j-1}\left|q^{t} \nabla_{z}^{\mathrm{pos}}\left(\mathcal{L}^{t}\right)\right|}{\sum_{t=0}^{j-1}\left|r^{t} \nabla_{z}^{\mathrm{neg}}\left(\mathcal{L}^{t}\right)\right|}.
\end{aligned}
\end{equation}
The positive and negative gradients of the~$t$-th iteration are represented by~$\nabla_{z}^{\mathrm{pos}}\left(\mathcal{L}^{t}\right)$ and~$\nabla_{z}^{\mathrm{neg}}\left(\mathcal{L}^{t}\right)$ respectively. These gradients are calculated based on the output of classifier~$z$ in the~$t$-th iteration. The weights of positive and negative gradients for the~$t$-th iteration are given by~$q^{t}$ and~$r^{t}$. The cumulative ratio of positive and negative gradients for the~$j$-th iteration is defined by~${g^{j}}$.
The EQL v2 balances the training process of each class independently by using the positive and negative gradients of the classifier's output to weight long-tailed samples. It has been shown to improve the performance of these samples on different baseline models. The idea has been extended to the one-stage object detection task by EFL~\cite{li2022equalized}, which uses two category-relevant factors to rebalance the loss contribution of positive and negative samples for each category. In general, the long-tailed distribution problem is better addressed by integrating the historical gradient of the long tail through the loss weighting method, either directly or indirectly. Additionally, in anomaly detection, abnormal samples have different gradients from normal samples. Kwon et al.~\cite{kwon2020backpropagated} use the gradient loss as a regularization term in the overall loss, which emphasizes the historical gradient of abnormal samples. The specific calculation process is as follows:
\begin{equation}
\begin{aligned}
\nabla_{x_{avg\_i}}^{j-1}=\frac{1}{(j-1)} \sum_{t=0}^{j-1} \nabla_{x_{i}}^{t}, 
\label{grad_avg}
\end{aligned}
\end{equation}
\begin{equation}
\begin{aligned}
\mathcal{L}_{\text {grad }}=-\underset{i}{\mathbb{E}}\left[{cos\_similarity}\left(\nabla_{x_{avg\_i}}^{j-1}, \nabla_{x_{i}}^{j}\right)\right].
\label{grad_cos}
\end{aligned}
\end{equation}
Eq.~\eqref{grad_cos} represents the cosine similarity between the average gradient~$\nabla_{x_{avg_i}}^{j-1}$ of~$(j-1)$-th iteration and the gradient~$\nabla_{x_{i}}^{j}$ of~$j$-th iteration, where~$i$ is the index of the layer in the decoder.
}

\noindent\textbf{Loss Values.} The value of the loss may depict the training stability, learning difficulty, and other beneficial optimization signals. {Many historical parameters are used in the loss calculation and update process, therefore loss itself could be regarded as a historical collection. Some studies have used the unique properties of the loss to identify anomalous samples~\cite{chu2020neural} or biased samples~\cite{jiang2021delving,huang2019o2u,xu2021small,yue2022ctrl,lazarou2021iterative,xia2021sample}. For instance, Chu et al.~\cite{chu2020neural} use loss profiles as an informative cue for detecting anomalies, while~\cite{jiang2021delving} and \cite{huang2019o2u } record the loss curve of each sample as an additional attribute to help identify the biased samples. Notably,~\cite{huang2019o2u } proposes O2U-Net to transfer the network from overfitting to underfitting cyclically by adjusting the hyper-parameters. O2U-Net splits the training process into three parts: pre-training, cyclical training, and training on clean data. Inside, cyclical training records the historical loss of each sample, then calculates and ranks every sample's normalized average loss to recognize the noisy data with a higher score. Later, some works introduce the value of historical loss to determine whether the data is with noisy label~\cite{ xu2021small,yue2022ctrl,lazarou2021iterative,xia2021sample }. To achieve a better performance of cleaning samples, Xu et al.~\cite{xu2021small} also use the historical loss distribution to handle the issue of varying scales of loss distributions across different training periods. 
\ylfChecked{Although average loss has shown promising results~\cite{huang2019o2u}, the existence of outliers in the training data can result in deviations when cleaning the data.}
Hence Xia et al.~\cite{xia2021sample} construct two more robust mean estimators to reduce the uncertainty of small-loss and large-loss examples separately. Furthermore, Yue et al.~\cite{yue2022ctrl} explore the algorithm for cleaning the labels based on historical loss, efficiently using historical information to obtain high-quality and noise-free data.}

\subsection{Aspect of Functional Part~\label{sec_3.2}}
{In Section~\ref{sec_3.1}, the background and details of the historical type are thoroughly discussed. Next, we elaborate on the appropriate functional parts in this subsection.

\noindent\textbf{Data.} 
\ylfChecked{The functional part of various data can be functionally categorized into four aspects. The first aspect addresses the problems of small numbers of valid positive and negative pairs and the imbalance of sample categories during training.} Adopting the memory bank or related variants~\cite{li2021spatial,zhuang2019local,sprechmann2018memory,wang2020cross, liu2022densely,wang2022contrastive,deng2022insclr,yang2022online,deng2021variational,li2020unsupervised,feng2021exploring, liu2022memory, wen2021false, liu2022learning, liu2021noise,ortego2021multi, jin2021mining, yang2022cross,wu2018improving,wu2018unsupervised, misra2020self, tian2020contrastive, yin2021instance, dai2021dual,zhou2022regional,xiao2017joint,alonso2021semi,lee2021weakly,li2022recurrent,liu2021one,yang2018learning,yang2019visual,zheng2021group} stores the past high-level instance embeddings to augment/formulate the critical training pairs to save feature computing budgets. 
It can be regarded as an efficient augmentation from the aspect of training data in the feature space, yet not the input (image) space. {Moreover, Zhang et al.~\cite{zhang2021rethinking} take hard negative mining with a rough similarity margin instead of easy negatives to improve the quality of the memory bank. Ge et al.~\cite{ge2020self} aim to learn features by taking all available data from source and target domains encoded using the hybrid memory module.} 
The second aspect involves gathering the feature statistics of the same class and dispersing the feature statistics of different classes. It is mainly achieved by reducing the intra-class distance and increasing the inter-class distance through the historical category representation. Center loss~\cite{wen2016discriminative} maintains the data center of each category by gradually accumulating the historical instance embeddings, and the data centers are utilized for learning the intra-class compactness. \ylfChecked{Later, diverse improvements~\cite{ he2018triplet,li2019angular,min2020adversarial,zhao2020deep,narayan20193c,chen2021free,li2021frequency, keshari2020generalized, jing2021cross} are proposed based on the center loss to address its limitations, thus facilitating a faster reduction of intra-class distance and expansion of inter-class distance.}
For the third aspect, some studies employ the unique properties of the loss to clean the dataset. They~\cite{chu2020neural,jiang2021delving,huang2019o2u,xu2021small,yue2022ctrl,lazarou2021iterative,xia2021sample} leverage the loss curve to reweight the loss function so as to identify anomalous/biased samples, and ultimately improve data cleanliness. 
Last, the fourth aspect focuses on employing historical input to perform data augmentation. Yang et al.~\cite{yang2022recursivemix} zoom and paste the historical input onto the new round of training images to make mixed-sample data augmentation.}

\noindent\textbf{Model.}
Existing works mainly focus on using parameters and gradients in the past model to improve the model's performance.
First, the previous model parameters are collected to build the teacher models~\cite{caron2021emerging,chen2020improved,he2020momentum,feng2021temporal,chen2021empirical,grill2020bootstrap,yuan2020revisiting,li2019learning,yang2019snapshot, furlanello2018born, tarvainen2017mean, chen2020big, tomasev2022pushing, mitrovic2020representation, van2021unsupervised, xie2021propagate, ayush2021geography, li2020prototypical,wang2022efficient,wang2022learn,wang2022self,cai2022semi, wang2022unsupervised, shi2018transductive,cai2021exponential, zhang2023robust,wang2021multi, kim2022pose, yu2019uncertainty, meng2021spatial, wang2021self,wang2022semi,liu2022perturbed, cui2019semi, perone2018deep,zhang2022semi, xu2021end, yang2021interactive , liu2021unbiased, liu2022unbiased, cai2019exploring, deng2021unbiased, xie2021detco,chen2021transferrable,luo2021coco,el2021training} that can offer effective supervisions. As a consequence, the performance of the student model is significantly boosted. 
In addition, model parameters are also performed by ensemble learning through averaging the model weights~\cite{zhang2020swa,athiwaratkun2018there,izmailov2018averaging,yang2019swalp,cha2021swad,guo2022stochastic,garipov2018loss,maddox2019simple,von2020neural} or predictions~\cite{huang2017snapshot,chen2017checkpoint,chen2018short,ilg2018uncertainty,poggi2020uncertainty,wolterink2018automatic,yu2019deep,liu2020loss,xu2020multi,gong2022uncertainty} during inference. The model weight is mainly constructed by unitary ensemble architecture in inference. \ylfChecked{For example, Zhang et al.~\cite{ zhang2020swa} average checkpoints from multiple training epochs trained with cyclical learning rates.}
\ylfChecked{Another way to use historical model parameters is to process model predictions, which emphasizes the exploitation of ensemble results from various historical models during inference.}
Both~\cite{ chen2017checkpoint} and~\cite{huang2017snapshot} employ a collection of model snapshots (or checkpoints) to assess each test sample and then combine the predictions as the final result. 
\ylfChecked{Batch Normalization~\cite{ioffe2015batch} and its variations~\cite{ioffe2017batch, li2016revisiting, salimans2016weight, arpit2016normalization} introduce some alternate applications, where the feature statistics are collected during training and subsequently used during testing.}
\ylfChecked{In contrast to using historical model parameters, optimization algorithms~\cite{liu2019variance,loshchilov2017decoupled,ding2019adaptive,kingma2014adam,sutskever2013importance,zeiler2012adadelta,duchi2011adaptive, dozat2016incorporating, chen2018closing, heo2020adamp, xie2022adan} perform adaptive gradient descent based on the historical gradient statistics, successfully speeding up the training convergence.}
\noindent\textbf{Loss.} The historical predictions~\cite{laine2016temporal,yang2022recursivemix, yao2020deep,kim2021self,nguyen2019self, shen2022self,chen2020semi}, feature embeddings~\cite{li2021spatial,zhuang2019local,sprechmann2018memory, wang2020cross, deng2022insclr, yang2022online, deng2021variational, li2020unsupervised,feng2021exploring, liu2022memory,xiao2017joint,alonso2021semi,lee2021weakly,li2022recurrent,liu2021one,yang2018learning, yang2019visual,liu2022densely,wang2022contrastive,wen2021false,liu2022learning,liu2021noise,ortego2021multi,jin2021mining,yang2022cross,zhong2019invariance,wu2018improving, wu2018unsupervised, misra2020self, tian2020contrastive, yin2021instance, dai2021dual,ko2021learning,zhou2022regional,zheng2021group,
wen2016discriminative,he2018triplet,li2019angular,min2020adversarial,zhao2020deep,narayan20193c,chen2021free,li2021frequency, keshari2020generalized, jing2021cross,fan2018associating,liu2020leveraging
}, gradients of various level features~\cite{tan2021equalization,li2022equalized,kwon2020backpropagated} and generated soft labels from self-constructed teachers~\cite{caron2021emerging,he2020momentum,chen2021empirical,feng2021temporal,grill2020bootstrap,chen2020improved, el2021training ,chen2021transferrable ,luo2021coco,chen2020big,van2021unsupervised,xie2021propagate,li2020prototypical,ayush2021geography,mitrovic2020representation,cai2019exploring, deng2021unbiased, xie2021detco,tomasev2022pushing,
yuan2020revisiting,li2019learning,yang2019snapshot,furlanello2018born,wang2022efficient,wang2022learn,wang2022self,
tarvainen2017mean,cai2022semi,wang2022unsupervised,shi2018transductive,wang2021multi, kim2022pose, yu2019uncertainty, meng2021spatial, wang2021self ,wang2022semi,liu2022perturbed, cui2019semi, perone2018deep,zhang2022semi, xu2021end, yang2021interactive, liu2021unbiased, liu2022unbiased, cai2021exponential, zhang2023robust
} \ylfChecked{can both contribute to }the loss objectives, providing meaningful and promising learning signals. {Besides, some methods directly use the value of loss itself to determine whether the sample is abnormal or noisy~\cite{chu2020neural,jiang2021delving,huang2019o2u,xu2021small,yue2022ctrl,lazarou2021iterative,xia2021sample}. For example,~\cite{huang2019o2u} not only record the loss curve of each sample but also average the loss to recognize the higher score noisy data.
\ylfChecked{In summary, collecting historical information to construct the loss or using the loss function itself as historical information represents an indirect or direct approach to leveraging past knowledge functionally.}
}

\subsection{Aspect of Storage Form~\label{sec_3.3}}
{In Section~\ref{sec_3.2}, we outline the specific functional parts of historical learning. This section covers the storage form of historical information. The various components of history learning are either recorded as Discrete Elements in the medium to long term, or they are transitory and disappear upon updating the Moving Average.}

\noindent\textbf{Discrete Elements (DEs)}: {DEs are stored in various ways over a medium to long term. The existing literature primarily stores DE in four forms. Memory bank or queue stores and updates cross-batch DEs by creating a container with limited capacity. The instance embeddings~\cite{li2021spatial,zhuang2019local,sprechmann2018memory,wang2020cross,liu2022densely,wang2022contrastive,deng2022insclr,yang2022online,deng2021variational,li2020unsupervised,feng2021exploring, liu2022memory, wen2021false, liu2022learning, liu2021noise,ortego2021multi,jin2021mining,yang2022cross,wu2018unsupervised, misra2020self, tian2020contrastive, yin2021instance, dai2021dual,he2020momentum, chen2020improved, el2021training,chen2021transferrable,zhou2022regional,luo2021coco,li2020prototypical,ayush2021geography,xie2021propagate,van2021unsupervised,chen2020big,xiao2017joint,alonso2021semi,lee2021weakly,li2022recurrent,liu2021one,wang2022semi,yang2018learning, yang2019visual,xie2021detco,zheng2021group} are added to the memory bank or queue for further use. These past elements are recorded and will be replaced based on the first-in, first-out principle when the memory bank or queue reaches its capacity.
The iterative teacher model maintains historical DEs~\cite{wang2022efficient,wang2022learn,yuan2020revisiting,yang2019snapshot,furlanello2018born,li2019learning,feng2021temporal}. Feng et al.~\cite{feng2021temporal} perform teacher learning by incorporating multiple historical model checkpoints in a parallel manner. Yuan et al.~\cite{yuan2020revisiting} suggest training the model using a single pretrained teacher model.~\cite{yang2019snapshot,furlanello2018born,li2019learning} use iterative teacher-student learning by employing multiple checkpoints as the teacher model in sequence. \ylfChecked{Generally, the iterative teacher model retains and updates previous DEs and subsequently provides supervision to the current student model during the training process.}
In contrast, checkpoints or snapshots from the learning history can be directly saved to create robust ensemble results~\cite{huang2017snapshot,chen2017checkpoint,chen2018short,ilg2018uncertainty,poggi2020uncertainty,wolterink2018automatic,yu2019deep,liu2020loss,xu2020multi,gong2022uncertainty}. 
Finally, other DEs such as prediction~\cite{kim2021self,shen2022self,yang2022recursivemix}, gradient~\cite{li2022equalized,tan2021equalization}, and loss~\cite{chu2020neural,jiang2021delving,huang2019o2u,xu2021small,yue2022ctrl,lazarou2021iterative,xia2021sample} \ylfChecked{may incorporate memory by storing their respective historical values for use in various operations.}

\noindent\textbf{Moving Average (MA).} There are typically two common forms of MA in the literature:

(1) Exponential Moving Average (EMA) is the most commonly used operator. {EMA calculates the local mean of a variable by considering its historical values over time. It typically updates momentum in predictions of probabilities~\cite{laine2016temporal,yao2020deep,nguyen2019self}, feature representations~\cite{li2021spatial,zhuang2019local,zhong2019invariance,wu2018improving,zhou2022regional,wen2016discriminative,he2018triplet,li2019angular,min2020adversarial,zhao2020deep,narayan20193c,chen2021free,li2021frequency, keshari2020generalized, jing2021cross,fan2018associating,liu2020leveraging,zheng2021group}, feature statistics~\cite{ioffe2015batch,ioffe2017batch, li2016revisiting, salimans2016weight,arpit2016normalization}, network parameters~\cite{caron2021emerging,chen2020improved,he2020momentum,feng2021temporal,chen2021empirical,grill2020bootstrap,tarvainen2017mean,chen2020big,van2021unsupervised,xie2021propagate,li2020prototypical,ayush2021geography,mitrovic2020representation,tomasev2022pushing,cai2022semi, wang2022unsupervised, shi2018transductive,cai2021exponential, zhang2023robust,wang2022self,li2019learning,wang2021multi, kim2022pose, yu2019uncertainty, meng2021spatial, wang2021self,wang2022semi,liu2022perturbed, cui2019semi, perone2018deep,zhang2022semi, xu2021end, yang2021interactive , liu2021unbiased, liu2022unbiased, cai2019exploring, deng2021unbiased, xie2021detco,chen2021transferrable,luo2021coco,el2021training}, and model gradients~\cite{liu2019variance,loshchilov2017decoupled,ding2019adaptive,kingma2014adam,sutskever2013importance,zeiler2012adadelta,duchi2011adaptive, dozat2016incorporating, chen2018closing, heo2020adamp, xie2022adan}. The primary goal of combining past predictions with EMA is to provide soft targets that standardize the model and minimize the supervised learning objective. Batch Normalization~\cite{ioffe2015batch} and its variants use the mean and variance of historical statistics computed through EMA to standardize the data. Besides, leveraging historical feature representations~\cite{zhong2019invariance,wu2018improving,wen2016discriminative}, network parameters~\cite{caron2021emerging,chen2020improved,he2020momentum,feng2021temporal,chen2021empirical,grill2020bootstrap,tarvainen2017mean}, and gradients~\cite{liu2019variance,heo2020adamp, xie2022adan} by EMA give more weight and importance to the most recent data points while still tracking a portion of the history.}

{(2) Simple Moving Average (SMA) is a type of arithmetic average that is calculated by adding recent data and dividing the sum by the number of time periods. The works in~\cite{zhang2020swa,wang2022efficient,wang2022learn,athiwaratkun2018there,maddox2019simple,garipov2018loss,izmailov2018averaging,von2020neural,guo2022stochastic,cha2021swad,yang2019swalp,chen2020semi,kwon2020backpropagated} all use SMA to obtain the average multiple checkpoints, predictions or gradients of features during the learning process. SMA has been proven to result in wider optima and better generalization.
Zhang et al.~\cite{zhang2020swa} focus on using SMA after additional training. This approach maximizes information retention but increases additional training costs.
Other works use SMA in different ways, such as following the optimization trajectory~\cite{wang2022self, wang2022efficient,wang2022learn, maddox2019simple, garipov2018loss, guo2022stochastic}, processing training later~\cite{von2020neural}, sampling from a Gaussian distribution~\cite{izmailov2018averaging} or finding high-accuracy pathways between loss functions~\cite{athiwaratkun2018there}. Additionally, some studies use the average of loss to recognize the noisy data~\cite{huang2019o2u,xu2021small,yue2022ctrl,lazarou2021iterative,xia2021sample}. 
\ylfChecked{These techniques aim to decrease the training cost while seeking suboptimal solutions that are in close proximity to the optimal solution when averaged.}
}

\section{Discussion}

\subsection{Relation to Related Topics}

\noindent\textbf{Relation to Recurrent Network.} Recurrent neural networks (RNNs)~\cite{zaremba2014recurrent,hochreiter1997long}  are neural sequence models that have played an important role in sequential tasks like language modeling~\cite{mikolov2010recurrent}, speech recognition~\cite{graves2013speech} and machine translation~\cite{kalchbrenner2013recurrent,bahdanau2014neural}. A well-known algorithm called Back Propagation Through Time (BPTT)~\cite{werbos1990backpropagation,guo2013backpropagation} is adopted to update weights in RNNs. When training, RNNs are expanded over time and have shared model parameters at each time step. Note that the ``time step'' here and  the ``time'' of BPTT in RNNs is quite different from the concept of time in learning history--they are in fact defined in the perspective of sequential dimension, whilst the time in learning history refers to a complete training iteration (or epoch). Therefore, the information from all expanded steps of RNNs can be together viewed as a set of historical knowledge at the same moment during learning. The hidden states of RNNs are sometimes called ``memory''~\cite{hochreiter1997long}, where the memory is actually the intermediate feature map across the input context, instead of the historical representations in our topic.

\noindent\textbf{Relation to Memory Network.} The series of Memory Network aims at reasoning with a external memory component~\cite{weston2014memory,graves2014neural,sukhbaatar2015end,chandar2016hierarchical} and is widely applied in Question Answering (QA) tasks.
The external memory is usually designed to be read and written to, which effectively maintains the long-term information. Different from the stored past feature representations during training history, the memory matrix acts as a global knowledge base that directly supports the inference over the entire sequence. It is more similar to the hidden states of RNNs but with a global long-term mechanism and shared read/write functions.

\noindent\textbf{Relation to Ensemble learning.} {Ensemble learning is an umbrella term for methods that combine multiple inducers to make a decision~\cite{sagi2018ensemble}. 
The utilization of ensemble learning methods is prevalent in historical learning, however, ensemble learning doesn't necessarily require historical information. These two concepts are separate, but can be combined.
Ensemble learning has advanced significantly in the field of machine learning~\cite{hansen1990neural},~\cite{dietterich2000ensemble},~\cite{caruana2004ensemble}. To achieve the best ensemble effect at the lowest cost, some work investigates how various hyperparameters~\cite{saikia2020optimized, wenzel2020hyperparameter }, architectures~\cite{zaidi2021neural, gontijo2021no }, strategies~\cite{levesque2016bayesian, wenzel2020hyperparameter, fei2022meta}, and distribution shift~\cite{ovadia2019can,mustafa2020deep} affect ensemble models. Ensemble learning primarily makes use of historical information by ensembling the historical results~\cite{huang2017snapshot, chen2017checkpoint} and parameters~\cite{wang2022efficient,wang2022learn} to enhance the model’s performance during training and inference. 
However, two exceptional cases cannot be considered as part of historical learning. Some works ensemble multiple classifiers~\cite{sensoy2018evidential,lakshminarayanan2017simple} and models~\cite{ruiz2019adaptative, yang2020resolution, wang2020glance, park2019feed,radosavovic2018data} into a single framework through the straightforward training process, which doesn't involve the integration, utilization, and updating of classifiers or branches during iterations or epochs. The past parameters, models, or other information are only temporarily stored and utilized, rather than being accumulated and transferred to the subsequent iteration or epoch, as is typical in historical information. This can be viewed as a brief integration of information before and after the training or reasoning process, not as an integration of historical information.
{Likewise, the outcomes of combining independent or parallel training models cannot be considered as historical learning~\cite{solovyev2021weighted, lakshminarayanan2017simple, wortsman2022model, wortsman2022robust,gupta2020stochastic}. For example, WiSE-FT~\cite{ wortsman2022robust } average the weights of zero-shot and fine-tuned models, and Model Soup~\cite{ wortsman2022model } average the weights of independently fine-tuned models. There is no historical connection between the results, and thus, there is no iterative storage, utilization, and updating. Unlike previous research~\cite{huang2017snapshot, chen2017checkpoint}, the current results are not enhanced by utilizing previous results and historical data, but by combining other results of independent or parallel training models.}

\noindent{\textbf{Relation to Reinforcement Learning.}} 
{Reinforcement learning (RL) is a subfield of machine learning that deals with learning how to control a system to maximize a long-term target value~\cite{szepesvari2010algorithms}. The concept of an experience replay buffer was first introduced by Lin et al.~\cite{lin1992self} and has since seen significant development and advancements in the field of RL~\cite{mnih2015human, lillicrap2015continuous, schaul2015prioritized, wang2016sample, andrychowicz2017hindsight}. The buffer is initialized first and network training is conducted once the number of data stored in the buffer exceeds a certain threshold. It has been extended to various RL tasks~\cite{haarnoja2018soft, savinov2018episodic, pong2019skew, bohmer2019exploration} and improved with related variants~\cite{schaul2015prioritized, horgan2018distributed, badia2020never}. {The experience replay buffer and memory bank~\cite{wang2020cross} have similarities and differences. The similarities lie in the fact that they both store information for training; the differences lie in that the target stored in the replay buffer can exist independently of the network~\cite{zhang2017deeper}, being related only to the RL environment and essentially serving as training data for RL. However, memory bank usually stores representations outputted by the network in its learning history.}
Thus, the experience replay buffer is typically not considered as historical learning. Some studies use ensembling different information~\cite{wiering2008ensemble, pathak2019self, shyam2019model, chua2018deep, henaff2019explicit} to make better decisions, but these works cannot be considered as historical learning since they fail to utilize historical information during the optimization process of each individual model.}

\subsection{Future Challenges}
\noindent\textbf{Advanced/Downstream Vision Tasks.} 
Most existing methods that utilize historical statistics focus on image classification and metric/representation learning, while giving little attention to downstream vision tasks such as visual tracking~\cite{ma2015hierarchical, bertinetto2016fully, wang2018sint++ , danelljan2015convolutional}, object detection~\cite{ren2015faster,lin2017focal,tian2019fcos,li2020Generalized}, and segmentation~\cite{noh2015learning,long2015fully,he2017mask,tian2020conditional,wang2020solo,wang2020solov2}. Although the potential benefits of historical learning in these tasks are significant, it has not been widely adopted. Semantically rich and task-specific elements or results can be generated during training and inference in many vision tasks, such as bounding boxes, keypoints, uncertainty, or other statistics. These pieces of information could be combined with learning history, such as improving the effectiveness of object detection based on the historical bounding boxes or improving semantic segmentation performance by leveraging the historical uncertainty. In short, incorporating task-specific elements with learning history can probably lead to a stronger model for these downstream vision tasks.
}

\noindent\textbf{Rarely Explored Directions.} Although several types of historical information (e.g., intermediate feature representations and model parameters) have been extensively explored, there is a considerably large room for many other directions--the historical predictions, inputs, labels, and gradients to the feature maps. These directions, along with the potential combinations of all types, lack deep discussion and further exploitation currently. 
Typically, the field of gradients and loss requires further investigation. Currently, the emphasis is mainly on utilizing historical gradients of model parameters through optimizers, but there is room for more effective methods. Tuluptceva et al.~\cite{tuluptceva2020perceptual} leverage the historical L2-norm of unique model parameters' derivatives to balance losses. However, the potential of historical gradients of all levels has yet to be fully explored. Further research is needed to discover new ways to use historical gradients of features, such as incorporating information on different categories. The same is true for historical loss, where most studies have only focused on using it for data cleaning during training. Although some works have utilized historical loss for effective data cleaning~\cite{xu2021small, huang2019o2u, xia2021sample, yue2022ctrl}, there is scope for improvement by combining it with other feature representations, such as using historical loss to determine the weight of feature maps between channels.

\noindent\textbf{Extensions to Other Fields.} Currently, it is observed that the majority of existing historical learning approaches focus on computer vision (CV) tasks. However, natural language processing (NLP) tasks~\cite{devlin2018bert,radford2018improving,radford2019language,brown2020language,ouyang2022training} have yet to heavily incorporate historical mechanisms\footnote{Here, we exclude advanced optimizers that use historical gradients, which are applicable to all fields.}. This presents a promising opportunity to adapt historical learning mechanisms to NLP. Nevertheless, several challenges arise when extending similar ideas from CV to NLP tasks, mainly because the data format of language differs significantly from that of images. For example, while images can be augmented with flexible operations such as mix, crop, and resize, language is the opposite. The flexibility of operations in CV make it easier for models to incorporate historical information. Therefore, the effective utilization of historical learning mechanisms in NLP is still an open area that has great potentials to be explored and developed.

\noindent\textbf{Learning Patterns in Learning History.} We suspect that the learning trajectory may have meaningful patterns to guide efficient and robust training of the original models. It can be possible to train, e.g., a meta network~\cite{munkhdalai2017meta} for discovering such patterns through the recorded historical variables, or treat them as meta data~\cite{vanschoren2018meta} to learn new tasks much faster than otherwise possible.  

\section{Conclusion}
In this paper, we systematically formulate the topic ``Historical Learning: Learning Models with Learning History'' by reviewing the existing approaches from the detailed three perspectives (i.e., what, where, and how), discuss their relation to similar topics and show promising directions for future exploration. 
We hope this topic can serve as a meaningful, inspiring, and valuable design principle, and this survey can inspire the community to explore continuously.


%

\ifCLASSOPTIONcaptionsoff
  \newpage
\fi

{\small
	\bibliographystyle{IEEEtran}
	\bibliography{IEEEtran}
}
\end{document}